%% file: main.tex
\documentclass[runningheads]{llncs}

\usepackage{eccv}

\usepackage{eccvabbrv}

\usepackage{graphicx}
\usepackage{booktabs}
\usepackage[normalem]{ulem}
\usepackage{multirow}
\usepackage{cuted}
\usepackage{tabularray}
\usepackage{wrapfig}
\usepackage{arydshln}
\usepackage[table, dvipsnames]{xcolor}

\def\method{\emph{MATCH}}

\usepackage[accsupp]{axessibility}  

\usepackage[breaklinks]{hyperref}

\usepackage{orcidlink}

\begin{document}

\title{{MATCH}: Flow Matching for Multi-View Anomaly Detection} 

\titlerunning{{MATCH}: Flow Matching for Multi-View Anomaly Detection}

\author{Mathis Kruse \orcidlink{0009-0001-2789-7255} \and
Melissa Schween \orcidlink{0009-0009-1418-7906} \and
Bodo Rosenhahn \orcidlink{0000-0003-3861-1424}}

\authorrunning{M.~Kruse et al.}

\institute{
Institute for Information Processing, L3S, Leibniz University Hannover, Germany\
\email{\{kruse,schween,rosenhahn\}@tnt.uni-hannover.de}}

\maketitle

\input{sec/0_abstract}
\input{sec/1_intro}

\input{sec/2_related}

\input{sec/3_method}

\input{sec/4_experiments}
\input{sec/5_conclusion}

\section*{Acknowledgements}
{\small
This work was supported by the MWK of Lower Saxony within Hybrint (VWZN4219) and LCIS (VWZN4704), the Deutsche Forschungsgemeinschaft (DFG) under Germany’s Excellence Strategy within the Cluster of Excellence PhoenixD (EXC2122) and Quantum Frontiers (EXC2123), the European Union under grant agreement no. 101136006 – XTREME.
}

%
%
\bibliographystyle{splncs04}
\bibliography{main}

\input{sec/X_suppl}

\end{document}

%% file: sec/0_abstract.tex
\begin{abstract}
Detecting anomalies in industrial objects is an important topic for increasing production efficiency.
More complex objects often require the analysis of several view points, which has led to the field of multi-view anomaly detection.
We present~\method{}, the first multi-view anomaly detection method based on Flow Matching (FM).
With the ODE formulation of Flow Matching, we can estimate likelihoods and thereby derive an anomaly score to detect anomalies in multi-view image data at object, image, and pixel-level.
The architectural flexibility of FM models allows us to efficiently transform features of different spatial sizes to the normal distribution.
We evaluate thoroughly on the already established Real-IAD data set and are also the first to provide a comprehensive evaluation of popular anomaly detection methods for the MANTA-Tiny data set.
\method{} achieves state-of-the-art performance in both anomaly detection and segmentation, all while running on consumer-level hardware.
By omitting the costly divergence term needed for likelihood estimation, we ensure that~\method{} is usable in real-time production scenarios.
Lastly, several ablation studies are conducted to validate the methodological choices.
Code: \url{https://github.com/m-kruse98/MATCH}
\keywords{Anomaly Detection \and Flow Matching \and Multi-View Data}
\end{abstract}

%% file: sec/1_intro.tex
\section{Introduction}

Any production pipeline may inevitably produce some defective products, either through human error or machine failures.
Therefore, detecting anomalies has always been a key task in increasing the efficiency and robustness of any industrial manufacturing line.
However, anomalies can be very rare and are often impossible to predict beforehand, while there is a vast availability of defect-free data.
This has led to the birth of semi-supervised anomaly detection (AD), where models learn from normal data only.

\begin{figure}
    \centering
  \includegraphics[width=.8\linewidth]{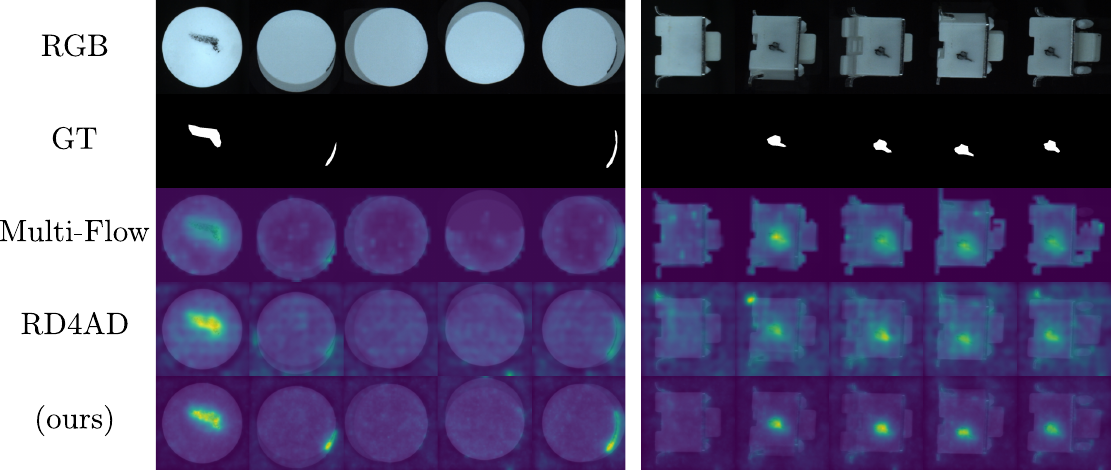}
    \caption{Qualitative comparison of Multi-Flow~\cite{MultiFlow}, Reverse Distillation (RD4AD)~\cite{rd}, and~\method{} (ours) on the \emph{white tablet} and \emph{short button} category of MANTA-Tiny. While Multi-Flow struggles with its foreground extraction and RD4AD detects some false positives,~\method{} most accurately segments all anomalies.}
    \label{fig:teaser_comparison}
\end{figure}

With the rise of deep learning, several popular AD benchmarks, along with well-performing methods, have been proposed~\cite{dataset:mvtec, dataset:visa}.
The bulk of AD research focuses on single-view images, where objects are usually rigidly aligned in the exact same spot in front of the camera.
However, as objects get more complex, a single camera may no longer suffice to detect all possible anomalies.
More sophisticated production lines and the ubiquitous usage of imaging systems have caused multi-view anomaly detection to gain more traction~\cite{dataset:realiad, dataset:pad}.
This lets the multi-view AD models also detect anomalies, that may not be covered by just a single camera.
The new setting of multi-view AD has not only spawned a variety of new data sets~\cite{dataset:realiad, dataset:manta, dataset:pad} but also many new AD methods~\cite{MultiFlow, splatpose, mvad_epipolar}.
An example of such a detection task is visualized in~\cref{fig:teaser_comparison}, where anomalies are only present in some views.

%
%
%

\begin{figure}[t]
    \centering
  \includegraphics[width=.85\linewidth]{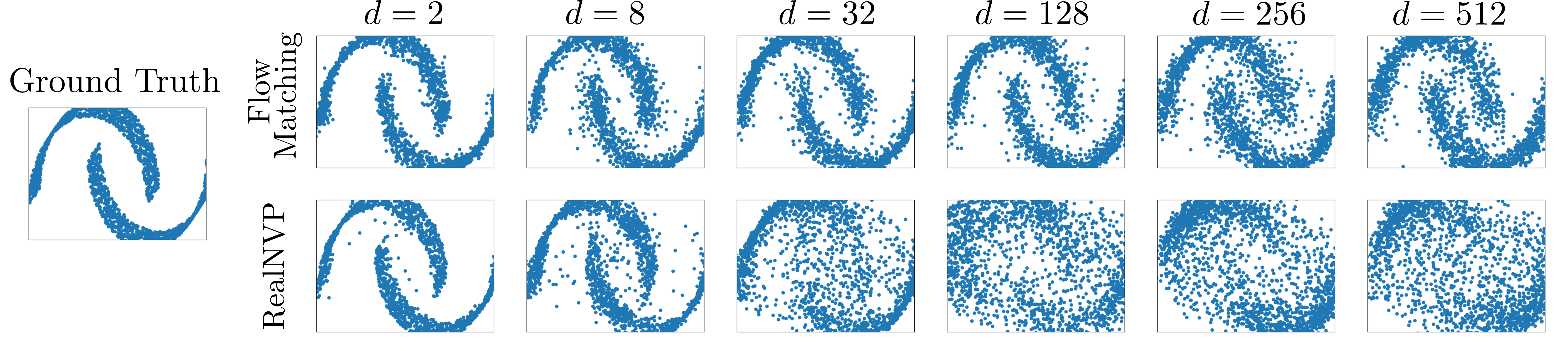}
    \caption{Sampling the "Two Moons" data set, with Flow Matching and RealNVP. Similar to AD problems, where small deviances need to be detected in a high dimensional feature space, we embed the two-dimensional moons in $d$-dimensional space. While the expressiveness of Flow Matching handles high dimensions well, the less flexible RealNVP architecture starts to struggle early on. Experimental details are in the appendix.}
    \label{fig:teaser_fm_vs_realnvp}
\end{figure}

Currently, a large variety of well-performing AD models use Normalizing Flows for estimating the likelihood of an image~\cite{differnet, csflow, fastflow, cflow, pyramidflow}.
However, these Normalizing Flow models, which are almost always chosen to be models based on RealNVP~\cite{realnvp}, may struggle to handle the high dimensionality of (multi-view) images.
Furthermore, their coupling block architecture limits them to a few transformation steps of expressivity.
A new generative modelling technique termed \textit{Flow Matching}~\cite{FM:seminal} has begun to gain interest very quickly.
Contrary to RealNVP, it can use more general network architectures and more easily represent even very high-dimensional data, while still functioning as both a generative model and a method for density estimation.
This is especially interesting for the task of multi-view AD.
Here, image data is usually very high dimensional, especially when different view points of an object need to be considered.
In AD, very small deviations in the data may already constitute an anomaly.
Thus, the decision boundary between normal and anomalous data may be a manifold of much smaller size, embedded within the high-dimensional data.
This makes flow matching better suited for multi-view anomaly detection.
A toy example, which visualizes the effect of learning a high-dimensional distribution, is shown in~\cref{fig:teaser_fm_vs_realnvp}.
Clearly, Flow Matching outperforms the inflexible RealNVP model, as it captures the underlying distribution even when it is embedded in higher dimensions.

Therefore, we propose \textbf{M}ulti-view \textbf{A}nomaly Detection using Flow Ma\textbf{tch}ing, abbreviated to~\method{}.
To the best of our knowledge, it is the first Flow Matching-based solution to multi-view AD.
We leverage the capabilities in expressiveness to process feature maps of different spatial resolutions.
Adding onto that, a few key simplifications are proposed to simplify the likelihood estimation and achieve new state-of-the-art results for both detecting and segmenting anomalies in popular multi-view AD benchmarks.
All the while, our model still remains fast to use, reaching $18.77$ FPS on consumer-level GPUs.

We summarize our contributions as follows:
\begin{itemize}
    \item We propose a Flow Matching-based multi-view anomaly detection model, outperforming previous methods on both Real-IAD and MANTA-Tiny.
    \item We  propose to omit the divergence term during likelihood estimation, which increases model throughput and thereby real-world usability, while leaving the detection performance unharmed and the score meaningful.
    \item We are the first to thoroughly benchmark the MANTA-Tiny data set.
    \item We release all code upon acceptance.
\end{itemize}

%% file: sec/2_related.tex
\section{Related Work}\label{sec:related_work}

We provide an overview of (multi-view) anomaly detection, as well as recent developments in Normalizing Flows, including Flow Matching.

\subsection{Anomaly Detection}\label{sec:related_anomaly_detection}

\paragraph{Overview.}
In recent years, research on anomaly detection (AD) has seen lots of success, with most of the popular benchmark data sets, such as MVTec AD~\cite{dataset:mvtec} solved with almost perfect scores~\cite{patchcore, simplenet, efficientAD}. 
In almost all cases, AD is learned in a semi-supervised fashion, where only anomaly-free data is available during training, and one model is trained per object class.
Some newer research started incorporating data modalities other than RGB, like depth information or normal maps~\cite{dataset:mvtec3d, dataset:realiadd3, dataset:eyecandies}.
Besides focusing on structural properties of single objects, some benchmarks also include logical anomalies, multiple objects, occlusions, and vast varieties of normal patterns~\cite{dataset:mvtecloco, dataset:mvtec2}. 

Besides data sets, there exist lots of categories of AD methods. Firstly, \emph{Reconstruction-based} methods aim to reconstruct the image features through some bottleneck mechanism~\cite{ae_ad,dae,dae_2,dinomaly}.
Closely related are \emph{student-teacher networks}, which distil information from a (potentially frozen) teacher network into a student network~\cite{AST_rudolph, efficientAD, rd, rrd, destseg}.
Some works also \emph{incorporate anomalous information}, either real or synthetically generated, to cast the task into a supervised problem~\cite{draem, realnet, simplenet, DRA}. 
As large vision-language-models (VLM) become more potent, \emph{zero-shot AD} focuses on detecting anomalies without any prior knowledge~\cite{TowardsTrainingFree, AnomalyClip, AnomalyGpt, WINClip}.
Using one model for all classes within a data set, a task termed \emph{multi-class anomaly detection}, has also seen a rise in popularity~\cite{uniad, mambaad, dinomaly}.

Many methods, including ours, may be grouped into the field of \emph{density estimation}.
By learning the distribution of the normal training data, the likelihood of a sample is approximated and used as the anomaly score. 
Methods based on k-nearest Neighbors (kNN) or memory banks form one strand of research~\cite{padim, SPADE, patchcore, cfa}, but rely heavily on the choice of features~\cite{heckler_features}.
A more principled way of estimating the density is offered by Normalizing Flows (NF), which have seen wide application in AD~\cite{differnet, csflow, MultiFlow, fastflow, cflow, pyramidflow}.
Here, the network explicitly learns the likelihood of a given sample using maximum likelihood estimation.
A natural next step from NFs is Flow Matching~\cite{FM:seminal}, which is the core of our method as will be explained in~\Cref{sec:related_flows,sec:method}.


\paragraph{Multi-view Anomaly Detection.}
Mirroring real industrial scenarios with complex objects, a single camera may not suffice to capture all anomalies. 
Therefore, research on multi-view anomaly detection has started gaining traction, with the Real-IAD data set~\cite{dataset:realiad} serving as a new important benchmark. 
It shows each object from five different view points, with anomalies having to be detected both in an image-wise, as well as an aggregated object/sample-wise fashion.
First methods for this exact setting have been proposed already~\cite{MultiFlow}, with some of them focused on multi-class AD~\cite{MVAD}.

Fan \etal~ present MANTA, a data set which features a camera setup very similar to Real-IAD~\cite{dataset:manta}. 
MANTA covers a much higher variety of samples and categories, while also being magnitudes larger than anything before.
Therefore, the authors also released MANTA-Tiny, a subset with largely the same test split, which is more easily usable for research purposes and solely used in this paper next to Real-IAD.
To the best of our knowledge, we are the first to report comprehensive benchmark results on MANTA-Tiny with our experiments.

Another strand of research on multi-view AD uses more complex objects with dozens or up to hundreds of views per object~\cite{dataset:RAD, dataset:pad, PIAD}.
Here, approaches based on novel-view synthesis methods have seen the most success~\cite{dataset:pad, PIAD, cvpr2025_PACLD, splatpose}. 
This remains, however, limited to one inflexible 3D model of normal samples, which fails when there is a larger variability of objects.
This is something naturally addressed by most other anomaly detection methods, as they do not explicitly save one normal prototype to compare to.
Concretely, density estimation techniques like Flow Matching allow for learning more complex distributions of normal samples. 

There exist other variants of AD that intersect with multi-view/3D methods, such as finding one abnormal object out of many~\cite{odd_one_out}, leveraging 3D ground truth information~\cite{looking3d}, or dealing with complex lighting scenarios~\cite{PIAD, M2AD}.
All of these multi-view settings have significantly increased the difficulty in anomaly detection benchmarks and thereby inspired the research community.  

\subsection{Normalizing Flows}\label{sec:related_flows}

\paragraph{Preliminaries.} The concept of Normalizing Flows~\cite{FM:guide_and_code} (NF) is based in the change of variables formula, which forms a relationship between a source distribution $p$ (usually a standard Gaussian) and the target or data distribution $q$, which are both distributions over $\mathbb{R}^d$.
The NF model $\phi: \mathbb{R}^d \rightarrow \mathbb{R}^d$ learns the mapping between data samples $x \sim q$ and $p$ via
\begin{equation}\label{eq:change_of_variables}
    q\left(x\right) = p\left(\phi\left(x\right)\right) \left|\det \left[ \frac{\partial \phi}{\partial x}\left(x\right) \right] \right|,
\end{equation}
with $p\left(\cdot\right)$ and $q\left(\cdot\right)$ denoting the respective probability density functions (PDFs).
Most notably, the flow $\phi$, usually parametrized by a neural network, needs to be both invertible and have a Jacobian whose determinant is easy to compute.
This has lead to the creation of specifically designed networks, that retain these properties~\cite{NF_NICE, nf1, nf2, glow_openai}, with RealNVP being used the most in the anomaly detection literature for its ease of use~\cite{realnvp, differnet, MultiFlow}.

Regular NFs usually chain together several invertible transformations $\phi = \phi_n \circ \ldots \circ \phi_1$, limiting their expressivity by the number of chained flows $n$.
This is alleviated with Continuous Normalizing Flows (CNF), which aim to learn the change in probability mass at every step using an ODE formulation~\cite{neural_ODE}.
The flow model now mimics the velocity field $u:\mathbb{R}^d \times \left[0, 1\right] \rightarrow \mathbb{R}^d$, learning the transformation needed to move a sample from $p_0$ at time $t=0$ to $p_1 = q$ at time $t = 1$.
In practice, however, CNFs are notoriously costly to train, requiring solving the entire ODE for a single training step~\cite{neural_ODE, ffjord}.

\paragraph{Flow Matching.} With Flow Matching (FM)~\cite{FM:seminal} training of CNFs has been simplified greatly by constructing conditional probability paths guiding from noise $p$ to data $q$.
Formally, an ODE is defined by its time-dependent vector field $u_t$, which is solved by its respective \emph{flow} $\phi_t(x_0)$ as
\begin{equation} 
    \frac{d}{dt} \phi_t\left(x_0\right) = u_t\left(\phi_t\left(x_0\right)\right),
\end{equation}
where the flow solution aims to satisfy
\begin{equation}
    \phi_t\left(x_0\right) \sim p_t \;\text{with initial condition}\;\, \phi_0\left(x_0\right)\sim p_0.
\end{equation}
Thus, we aim to learn the vector field $u_t$, such that its flow $\phi_t$ generates the probability path from $p_0$ to $p_1$, where each $p_t$ is again a distribution over $\mathbb{R}^d$.
As shown by Lipman \etal~\cite{FM:seminal}, the marginal vector field $u_t\left(x_0\right)$ turns out to be equivalent to the expectation of the conditional vector field over the data distribution as 
\begin{equation}\label{eq:vector_field}
    u_t\left(x_0\right) = \mathbb{E}_{x_1 \sim q} \left[ u_t\left(x_0 | x_1\right)\right].
\end{equation}
Thereby, sampling data $x_1 \sim q$ and regressing against the conditional vector field $ u_t\left(x_0 | x_1\right)$ suffices for learning a valid flow.
This conditional vector field can easily be constructed given the available data, \eg using Gaussian probability paths.
This also results in the loss becoming a simple regression against the constructed conditional vector field.
Subsequent work has explored different probability paths and other formulations, resulting in models with strong generative capabilities~\cite{FM:OT, FM:latent}.

%% file: sec/3_method.tex
\section{Method}\label{sec:method}

\paragraph{Problem Statement.} During training, anomaly-free multi-view images of a certain object category are given. 
At inference, both anomalous and defect-free images appear, with the model having to detect the occurring anomalies.
An object is considered anomalous if there is an anomaly visible in any of its views.

\paragraph{Objective.} As motivated with~\cref{fig:teaser_fm_vs_realnvp}, the distribution of normal images is learned using FM, specifically the optimal transport conditional Flow Matching (OT-CFM) formulation~\cite{FM:OT}.
The loss for optimizing the parameters $\theta$ of the flow network $u_t^\theta$ is
\begin{equation}\label{eq:ot_cfm_loss}
    \mathcal{L}_{\text{CFM}}^{\text{OT}} = \mathbb{E}_{\begin{subarray}{c} x_0 \sim p_0, x_1 \sim q \\ t \sim \mathcal{U}[0,1] \\ \end{subarray}}\; || u_t^\theta \left(x_t\right) - \left(x_1 - x_0\right) ||^2,
\end{equation}
where $p_0$ is standard normal Gaussian noise, $q$ the data distribution and $x_t = (1-t) x_0  + t x_1$ an intermediate point along the flow.

Using the trained $u_t^\theta$ and any off-the-shelf ODE solver, we can use our algorithm in two ways.
Running the ODE from $t=0$ to $t=1$ transforms noise $x_0 \sim p$ into data samples $x_1 \sim q$ (generative direction). Reversing it, and integrating the ODE from $t=1$ to $t=0$ lets us move real data samples to the Gaussian space, where we are able to measure likelihoods (density estimation direction).
The likelihood of a sample $x$ will serve as a proxy for the anomaly score and may be calculated as
\begin{equation}\label{eq:log_likelihood_equation}
    \log p_1 \left( x_1 \right) = \log p_0 \left(x_0\right) - \int_0^1 \text{div}\left(u^\theta_t\right) \left(x_t\right) dt,
\end{equation}
where $p_0 \left(x_0\right)$ is the density of the resulting sample in Gaussian space and $\text{div}\left(u_t^\theta\right)$ the divergence of the vector field at the intermediate solutions $x_t \sim p_t$ along which we integrate.
In practice, this divergence term is estimated using methods such as Hutchinson's trace estimator~\cite{hutchinson} or randomized quasi monte-carlo (RQMC)~\cite{rqmc}.
Even then, estimating the divergence term would result in (at the very least) one backpropagation per ODE step, which would vastly increase the computational requirements.
This would leave the model unsuitable for deployment in real-time industrial production scenarios.
An ablation conducted in~\cref{sec:ablation_divergence} shows that omitting the divergence term greatly saves computation time and memory while still performing well for AD.

Thus, our anomaly score just becomes $S\left(x\right) = -\log p_0 \left(x_0\right),$ which in practice, calculated along the feature dimension, becomes 
\begin{equation}\label{eq:anomaly_score}
    S(x_1) = -\log p_0\left(x_0\right) = \frac{||x_0||^2}{2} + \frac{d}{2}\log\left(2\pi\right),
\end{equation}
where $x_0$ is the sample in Gaussian space, and $d$ the feature dimensionality. 
This score may be interpreted, up to the additive constant, as the squared Mahalanobis distance of the sample $x_0$ to the standard Gaussian $\mathcal{N}\left(0, I\right)$. This interpretation is used in many related papers on AD and normalizing flows~\cite{differnet, csflow, cflow}.

\subsection{Model Architecture}
Normalizing flows trained on raw images have been shown to fail in detecting anomalies~\cite{nf_kirichenko_ood}.
Further, Flow Matching, just as diffusion models, performs well on latent feature distributions~\cite{FM:latent}.
Therefore, the flow is instead learned on the latent distribution of a WideResNet~\cite{wideresnet} (pre-trained on ImageNet~\cite{imagenet}) which serves as a frozen feature extractor.

\begin{figure*}[t]
    \centering
    \centering
    \includegraphics[width=0.75\linewidth]{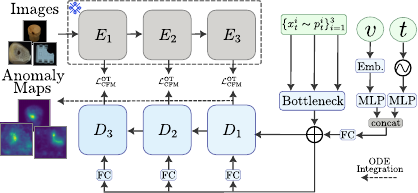}
    \caption{Overview of~\method. A frozen feature encoder $E_i$ is used for latent embeddings. With these, intermediate latent vectors $x_t^i$ are sampled, and, together with time $t$ and view $v$ information, fed through a bottleneck, and into the decoder layers $D_i$. The Flow Matching loss $\mathcal{L}_{\text{CFM}}^{\text{OT}}$ is applied and lets our model learn the trajectory from data space $q$ to Gaussian space $p$. Running the model with an ODE from $t=1$ to $0$ lets us produce anomaly maps, which segment any anomalous regions.}
    \label{fig:rd_architecture}
\end{figure*}

We take inspiration by the \emph{ReverseDistillation} (RD) architecture~\cite{rd}, with a high-level schematic of our adapted architecture shown in~\cref{fig:rd_architecture}.
A longer and completely in-depth graph of all layers and blocks can be found in the supplement in~\cref{sec:appendix_arch}.
Firstly, the data samples are extracted from the first three layers $E_i$ of the WideResNet as $\{x_1^i \sim q^i\}_{i=1}^3$, where $i$ denotes the current layer of the extractor.
To train, a time step $t \in \mathcal{U}\left(0,1\right)$ and noise of the same size as the latent features $\{x_0^i \sim p^i\}_{i=1}^3$ are sampled. 
Then, the intermediate features $\{x_t^i = (1-t) x_0^i + tx_1^i\}_{i=1}^3$ are constructed as a linear combination and fed through a bottleneck.
After that, several ResNet-style layers $D_i$ in reversed order are used to reconstruct features to the correct spatial dimension, similar to the architecture for RD4AD~\cite{rd}.

\textbf{Time and View Embedding.} 
To make the model usable for Flow Matching, we embed the time step information $t$ using a $1024$-dimensional sinusoidal positional encoding~\cite{vaswani2017attention} and a simple two-layer MLP with SiLU activation~\cite{silu}.
Similarly, the index of the current view $v$ is fed into a learned embedding layer as well as an MLP with the same dimensions.
Both view and time information are concatenated, passed through another fully connected (FC) layer with SiLU activation and added to the output of the bottleneck.
They also serve as input to the reconstruction layers $D_i$, with each layer having their own FC and SiLU layer for fine-tuning features to their needs.

\textbf{Bottleneck Network.}
The input features $\{x_t^i \in \mathbb{R}^{C_i \times H_i \times W_i}\}_{i=1}^3$ are fed into the bottleneck network, consisting of an alignment network and bottleneck blocks.
For the \emph{alignment}, the map at $i=3$ has the lowest feature resolution.
We thereby run the other maps through a down sampling using $3 \times 3$ convolutional layers with stride $2$, using one layer for $i=2$, and two for $i =1$.
The spatially aligned maps are then concatenated to produce one feature vector.
Each of the mentioned convolutional layer consist of its respective convolution followed by group normalization with $32$ groups, and ReLU activation.
Afterwards, three ResNet-style \emph{bottleneck} blocks~\cite{rd, resnet} (again with group normalization) further distil the feature representation, forcing the network to focus on the normal structures found in training only.
Residual connections are either directly implemented, or using a downsampling with another convolution.


\textbf{Decoder Blocks.}
The decoder consists of three layers $D_i$ each consisting of $d_i$ decoder blocks, that progressively upsample the bottleneck feature representation back to the original spatial dimensions, which mostly follows RD~\cite{rd, deconvnet}.
These decoder blocks use $1 \times 1$ convolutions followed by $3 \times 3$ convolutions with a stride of $2$ for upsampling~\cite{deconvnet}, group normalization, ReLU activation and a final $1 \times 1$ convolution.
Here, the time-view embedding is again added to the latent features.
Again,  the residual connection may be done either using an identity map or a single transposed convolutional layer for upsampling.

\textbf{Anomaly Scores.}
We produce three anomaly score maps $M_i \in \mathbb{R}^{H_i\times W_i}$ by calculating the anomaly score along the feature dimension of our decoded features as in~\cref{eq:anomaly_score}.
All of these maps are resized to the full spatial dimension of our images using bilinear interpolation.
For segmentation, we simply use the combination of all maps $M_\text{segment} = \sum_i M_i$ as our final segmentation map.
This includes both good segmentation of high-level defects, as well as the tiny low-level and hard-to-spot anomalies.

For detection, it has been shown that higher level representations are better suited to produce a single per-image anomaly score~\cite{nf_kirichenko_ood}.
Thereby, we use $M_{\text{detect}} = M_2 + M_3$ as the sum of only the last two anomaly maps, which mostly focuses on higher-level concepts while not being influenced too much by lower level deviations.
The final \emph{image-wise} anomaly score is then produced by taking the maximum value found in this detection map $M_\text{detect}$.
To produce a \emph{sample-wise} anomaly score, we again take a maximum across all anomaly scores produced for one multi-view instance.

%% file: sec/4_experiments.tex
\section{Experiments}\label{sec:experiments}

We evaluate our method on data sets Real-IAD~\cite{dataset:realiad} and MANTA-Tiny~\cite{dataset:manta}, while also reporting numbers for many competing state-of-the-art approaches.

\subsection{Data Set and Metrics}\label{sec:dataset_and_metrics}

We use both Real-IAD~\cite{dataset:realiad} and MANTA-Tiny~\cite{dataset:manta} for evaluating our method.
Real-IAD~\cite{dataset:realiad} is a large multi-view anomaly detection data set, made up of 30 different object classes taken from industrial production contexts, totalling roughly 100k normal and 50k anomalous images.
Defects such as scratches, missing parts or deformations are some of the most common categories of anomalies.

MANTA~\cite{dataset:manta} contains multi-view images of \emph{tiny objects}, with the different object classes spanning across the categories of agriculture, medicine, electronics, mechanics, and groceries.
With MANTA being about five times the size of Real-IAD, a smaller benchmark called MANTA-Tiny has been proposed, where images are resized and the amount of training instances is reduced to roughly match previous data sets and increase its popularity.
The test sets of both splits remain largely the same.
Depending on the category, defects may appear as misprints, stains, scratches, mold, or mildew, amongst many other possible faults.

For quantitative results, we report detection performance using the Area Under the Receiver Operating Curve (AUROC), which is independent of a specific threshold.
The AUROC will be reported both \emph{image-wise} (\ie each image is treated independently, named I-AUROC) and \emph{sample-wise}, where all images belonging to the same object are grouped together to form one single anomaly score (named S-AUROC).
Lastly, for segmenting the anomalies, we also report the AUROC in a \emph{pixel-wise} fashion (P-AUROC), as well as P-AP (average precision), and the more balanced P-AUPRO metric~\cite{dataset:mvtec}.

\begin{table}[t]
\centering
\caption{Both segmentation and image-wise detection metrics averaged across all classes of Real-IAD~\cite{dataset:realiad}. The best performing method is marked in \textbf{bold} with the runner-up \underline{underlined}. Our method is run $n=5$ times.}
\begin{tabular}{l||c|c|c|c|c}
\multicolumn{1}{c||}{} & \multicolumn{2}{c|}{Detection Metrics ($\uparrow$)} & \multicolumn{3}{c}{Segmentation Metrics ($\uparrow$)} \\\toprule
Method                            & I-AUROC        & S-AUROC        & P-AUROC        & P-AUPRO        & P-AP            \\\midrule
CFA \cite{cfa}                    & 71.44          & 77.53          & 95.58          & 79.05          & 13.63           \\
DSR       \cite{zavrtanik2022dsr} & 58.86          & 52.83          & 75.47          & 56.02          & 17.03           \\
EfficientAD \cite{efficientAD}    & 77.26          & 86.91          & 89.64          & 71.31          & 13.18           \\
FastFlow \cite{fastflow}          & 79.26          & 92.30          & 90.18          & 66.99          & 06.16           \\
Multi-Flow \cite{MultiFlow}       & \uline{90.27}  & \textbf{95.85} & 96.47          & 87.91          & 12.42           \\
PaDiM  \cite{padim}               & 89.27          & 91.33          & 94.22          & 75.33          & 06.73           \\
RD4AD \cite{rd}                   & 89.26          & 93.46          & \uline{98.81}  & \uline{92.70}  & \uline{32.65}   \\
SimpleNet \cite{simplenet}        & 86.37          & 94.89          & 93.90          & 78.87          & 18.36           \\
STFPM \cite{stfpm}                & 71.74          & 74.99          & 91.41          & 74.47          & 15.83           \\
\method (\emph{ours}) & \textbf{91.17} \scriptsize{$\pm 0.08$} & \uline{95.63}  \scriptsize{$\pm 0.09$} & \textbf{99.24} \scriptsize{$\pm 0.01$} & \textbf{94.76} \scriptsize{$\pm 0.03$} & \textbf{32.68} \scriptsize{$\pm 0.03$} \\     
\end{tabular}
\label{tab:mean_results_realiad}
\end{table}

\subsection{Implementation Details}\label{sec:implementation}

All experiments are run on a Linux machine with a RTX 3090 consumer GPU.
All metrics are calculated efficiently using the GPU-optimized ADEval library~\cite{ader}.
We resize all images to $256 \times 256$ pixels, resulting in extracted feature maps of sizes $(256 \times 64 \times 64)$, $(512\times 32 \times 32)$, and $(1024 \times 16 \times 16)$ using the WideResNet50.

For optimization we use the AdamW optimizer~\cite{adamw} with a learning rate of $1e{-4}$, momentum parameters $\beta_{1,2} = \left(0.9, 0.95\right)$, and weight decay set to $0.01$.
Contrary to most NF models, we do not need any clipping or scaling for the gradients~\cite{ardizzone_nf}.
The decoder layers $D_i$ consists of $3$, $4$ and $6$ decoder blocks respectively.
All convolutional layers work with a base width of $768$ dimensions.
We train on both data sets for $150$ epochs with a batch size of $8$.
For inference using the ODE, we use the simple Euler method with step sizes of $0.2$, resulting in $5$ forward passes per inference batch.
We will later make this choice more founded in an ablation in~\cref{sec:ablation_ode}.

\paragraph{Baselines.} A variety of state-of-the-art AD methods are reproduced using the Anomalib~\cite{anomalib} and ADer~\cite{ader} benchmarking libraries.
We chose them to be executed with the same consumer-level GPUs and setup as our method.
The baselines include CFA~\cite{cfa}, DSR~\cite{zavrtanik2022dsr}, EfficientAD~\cite{efficientAD}, FastFlow~\cite{fastflow}, Multi-Flow~\cite{MultiFlow}, PaDiM~\cite{padim}, ReverseDistillation (RD4AD)~\cite{rd}, SimpleNet~\cite{simplenet}, and STFPM~\cite{stfpm}.

\subsection{Anomaly Detection Results}\label{sec:ad_results}


We run~\method{} on both Real-IAD and MANTA-Tiny.
For Real-IAD, all metrics are reported in~\cref{tab:mean_results_realiad}.
Here, we outperform all competitors in most metrics by convincing margins, reaching an I-AUROC of $91.17$ for detection, and an P-AUPRO of $94.76$ for segmentation, all while being very stable with a standard deviation of $.08$ and $.03$ respectively.
Most notably, the direct flow-based competitor Multi-Flow, based on RealNVP, underperforms in detection by roughly one point, while~\method{} is far more precise at segmentation, as visible in the P-AUPRO.
This strengthens us in our claim that FM is better able to handle the high dimensionality of the feature maps, as it is still able to pinpoint the locations of the occurring anomalies.

As for the sample-wise detection (with S-AUROC), we equip all baselines with a simple max-aggregation, just as used by \method, to calculate their scores.
We achieve an S-AUROC of $95.63$ on Real-IAD. This score only lacks slightly behind the most direct competitor Multi-Flow, which is also optimized for multi-view AD, by $.2$ points.
On all other metrics and tasks, we outperform the baselines, especially in segmentation tasks.

Segmentation results for the newly proposed and until now not thoroughly evaluated MANTA-Tiny data set are presented in~\cref{tab:aupro_manta,tab:pixel_auroc_manta}.
The most important results on segmentation are presented with the P-AUPRO in~\cref{tab:aupro_manta}.
Here, we achieve a P-AUPRO of $89.66$ across all categories, which outperforms all baselines.
We further note that classes containing a larger variance in their normal samples, such as the naturally produced agriculture and groceries categories, are far harder to solve than categories focusing on classical industrial contexts.
Also, the ReverseDistillation (RD4AD) model almost always comes in second, highlighting the effectiveness of the chosen architecture.
Still, the Flow Matching formulation managed to gain very strong improvements, with a P-AUPRO more than four points higher.

Only in the detection metric I-AUROC does~\method{} fall behind, as SimpleNet is able to get a score of $93.19$ while~\method{} reaches a competitive $90.66$.
We posit that this may be due to the much higher variance in defects found in MANTA-Tiny, as some tiny defects may be harder to translate into a high score for the entire image.
Handling this issue may be interesting for future work.
Lastly, we also evaluated our model (trained on MANTA-Tiny) on the full MANTA test split, which is roughly 70\% identical.
We observed no noticeable performance drops, as shown in the supplementary. This shows MANTA-Tiny to be a good real-world benchmark, that is worth exploring.

\begin{table*}[t]
\centering
\caption{Anomaly Segmentation results on MANTA-Tiny measured as \textbf{P-AUPRO} ($\uparrow$) and grouped by category. The best performing method is marked in \textbf{bold} with the runner-up \underline{underlined}. Our method is run $n=5$ times.}

\resizebox{\linewidth}{!}{
\begin{tabular}{l||c|c|c|c|c|c|c|c|c|c}
\multirow{2}{*}{Category}    &  CFA   & DSR   & Effic.AD & FastFlow & Multi-Flow & PaDiM & RD4AD    & SimpleNet & STFPM & \method          \\
& \cite{cfa} & \cite{zavrtanik2022dsr} & \cite{efficientAD} & \cite{fastflow} & \cite{MultiFlow} & \cite{padim} & \cite{rd} & \cite{simplenet} & \cite{stfpm} & \emph{(ours)} \\\midrule

Agriculture & 43.07  & 50.07 & 54.92                   & 48.33    & 69.41                  & 63.57 & \underline{69.74} & 59.92     & 48.22 & \textbf{75.52} \scriptsize{$\pm$ 0.05} \\
Electronics &  59.62 & 57.21 & 89.30                   & 72.18    & 84.35                  & 84.01 & \underline{92.79} & 84.71     & 56.91 & \textbf{93.93} \scriptsize{$\pm$ 0.02}  \\
Groceries   &  39.35 & 46.43 & \underline{66.72 }      & 43.81    & 63.28                  & 55.23 & 56.52             & 61.78     & 34.48 & \textbf{70.03} \scriptsize{$\pm$ 0.05} \\
Mechanics   &  71.35 & 65.34 & 87.72                   & 76.98    & 87.39                  & 86.98 & \underline{94.41} & 85.47     & 80.47 & \textbf{96.35} \scriptsize{$\pm$ 0.02} \\
Medicine    &  67.93 & 58.46 & 78.26                   & 75.66    & \underline{83.38 }     & 79.71 & 82.75             & 80.11     & 70.20 & \textbf{89.10} \scriptsize{$\pm$ 0.04} \\\midrule
MANTA-Tiny  &  61.64 & 58.26 & 80.83                   & 69.65    & 81.76                  & 79.43 & \underline{85.59} & 79.42     & 64.19 & \textbf{89.66} \scriptsize{$\pm$ 0.02} 
\end{tabular}
}
\label{tab:aupro_manta}
\end{table*}

\begin{table*}[t]
\centering
\caption{Anomaly Segmentation results on MANTA-Tiny measured as \textbf{Pixel-AUROC} ($\uparrow$) and grouped by category. The best performing method is marked in \textbf{bold} with the runner-up \underline{underlined}. Our method is run $n=5$ times.}
\resizebox{\linewidth}{!}{
\begin{tabular}{l||c|c|c|c|c|c|c|c|c|c}
\multirow{2}{*}{Category}    &  CFA   & DSR   & Effic.AD & FastFlow & Multi-Flow & PaDiM & RD4AD    & SimpleNet & STFPM & \method          \\
& \cite{cfa} & \cite{zavrtanik2022dsr} & \cite{efficientAD} & \cite{fastflow} & \cite{MultiFlow} & \cite{padim} & \cite{rd} & \cite{simplenet} & \cite{stfpm} & \emph{(ours)} \\\midrule
Agriculture & 70.84 & 63.73 & 84.50          & 71.67    & 84.92      & 83.36 & \underline{87.22} & 87.19              & 76.60  & \textbf{87.50} \scriptsize{$\pm$ 0.02}  \\
Electronics & 86.60 & 76.05 & 95.63          & 91.39    & 95.28      & 95.62 & \underline{98.28} & 96.49              & 83.18  & \textbf{98.47} \scriptsize{$\pm$ 0.00} \\
Groceries   & 73.48 & 64.92 & \textbf{86.78} & 71.05    & 85.58      & 75.25 & 83.26             & \underline{86.76}  & 67.36  & 85.92          \scriptsize{$\pm$ 0.04} \\
Mechanics   & 87.09 & 70.32 & 94.73          & 91.59    & 97.00      & 95.9  & \underline{98.28} & 96.41              & 92.70  & \textbf{98.55} \scriptsize{$\pm$ 0.00} \\
Medicine    & 82.85 & 67.54 & 90.17          & 89.62    & 91.70      & 91.62 & \underline{92.66} & 91.87              & 89.54  & \textbf{95.53} \scriptsize{$\pm$ 0.02} \\\midrule
MANTA-Tiny  & 83.16 & 70.20 & 92.21          & 87.35    & 93.07      & 91.85 & \underline{94.60} & 93.63              & 85.50  & \textbf{95.65} \scriptsize{$\pm$ 0.01} 
\end{tabular}
}
\label{tab:pixel_auroc_manta}
\end{table*}

\begin{figure*}[t]
    \centering
    \includegraphics[width=0.95\linewidth]{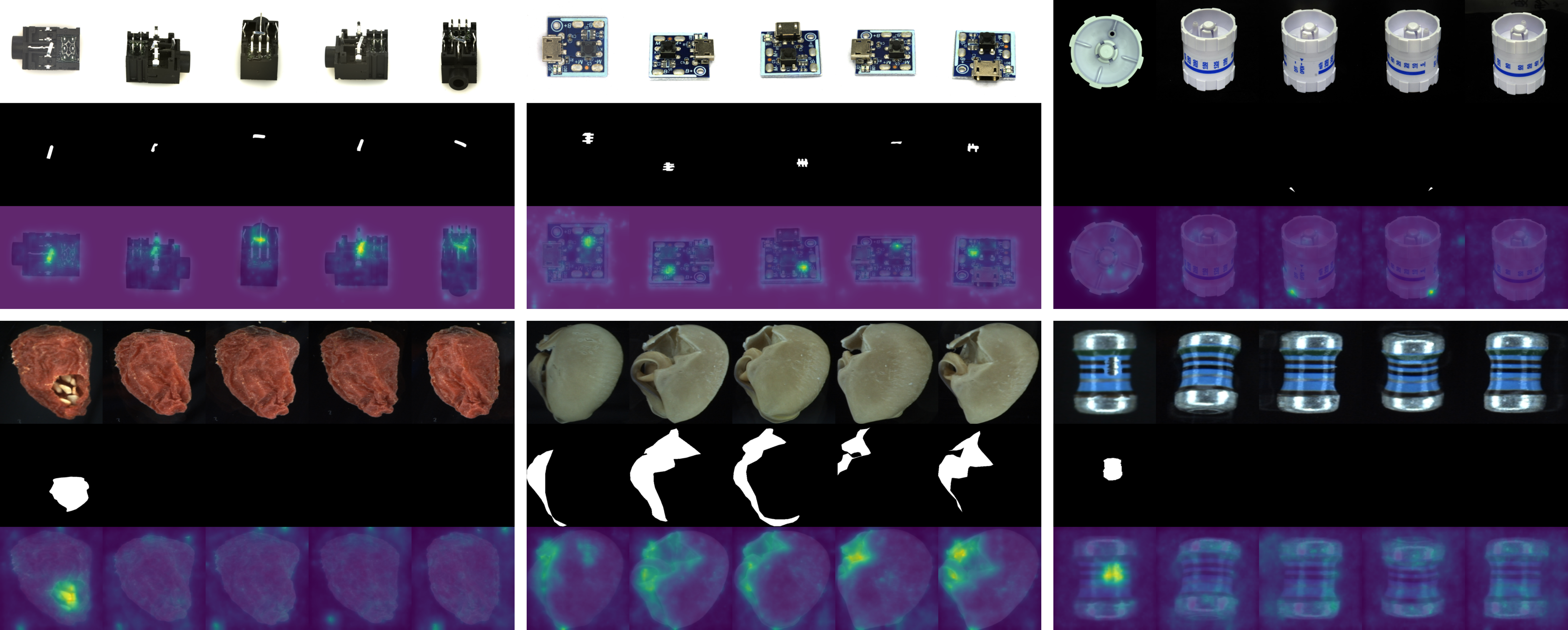}
    \caption{A selection of qualitative anomaly maps for classes audiojack, PCB, and regulator of Real-IAD (top row). The bottom row shows goji berries, soybeans, and wafer resistors of MANTA-Tiny. Pictured are the original RGB image, ground truth segmentation maps and the resulting anomaly maps overlayed on RGB.}
    \label{fig:qualitative_anomaly_maps}
\end{figure*}

For both of the mentioned benchmarks, we provide a variety of qualitative anomaly maps in~\cref{fig:qualitative_anomaly_maps}.
Here, the high segmentation quality of~\method{} becomes apparent, which is a clear advantage to preceding Normalizing Flow-based methods~\cite{csflow, MultiFlow}.
We also provide additional qualitative segmentations for every category of MANTA-Tiny, as well as an analysis of some failure cases in the supplement in~\cref{sec:appendix_qualitative}, together with the comprehensive results for all metrics on both data sets in~\cref{sec:appendix_comprehensive}.

\subsection{Ablation Studies}

\begin{wrapfigure}{r}{0.55\linewidth}
    \vspace{-\baselineskip}
    \centering
    \includegraphics[width=\linewidth]{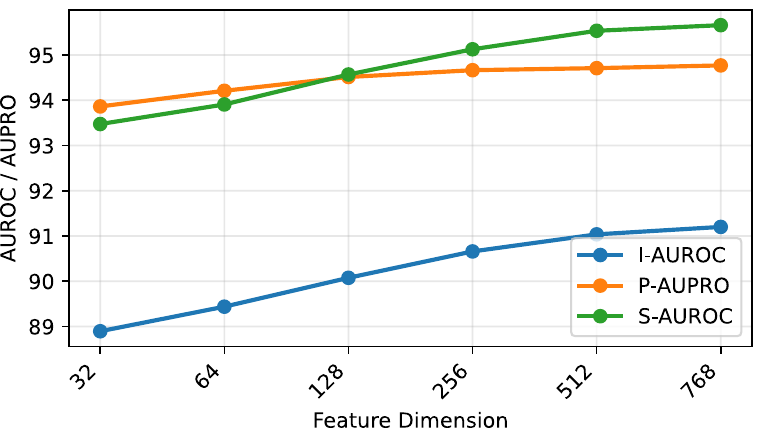}
    \caption{Varying the feature dimension for bottleneck and decoder from $32$ to $768$. All of the  metrics linearly rise with higher dimensions.}
    \label{fig:ablation_dimensionality}
    \vspace{-2\baselineskip}
\end{wrapfigure}

We conduct ablations to motivate design choices and provide further insight.
~\cref{fig:ablation_dimensionality} shows the impact of different base widths of our network. 
The AD performance being best around a width of $768$ confirms that strong and expressive (FM-based) models are most suited.
Also, matching the parameter count of RD4AD to~\method{}, RD4AD reaches an I-AUROC of $86.75$, confirming that gains stem from the FM formulation, and not just model capacity.
We do not consider higher widths, as we want~\method{} to remain usable on affordable consumer-level hardware.

\subsubsection{Impact of the Divergence Term.}\label{sec:ablation_divergence}

As described in~\cref{sec:method}, we omit the divergence term in~\cref{eq:log_likelihood_equation}, which is equal to the trace of the Jacobian matrix $\text{tr}[\partial_x u_t\left(x\right)]$, to increase the computational efficiency of our model.
There are several possible methods to estimate this divergence term.
Using Hutchinson's trace estimator, we may sample $M$ vectors $z \sim \mathcal{Z} = \mathcal{N}\left(0,I\right)$ and calculate
\begin{equation}
    \text{div}\left(u_t\right)\left(x\right) = \text{tr}\left[ \partial_x u_t\left(x\right) \right] = \mathbb{E}_{z \sim \mathcal{Z}} \; \text{tr}\left[z^T \partial_x u_t\left(x\right) z \right]
\end{equation}
as an unbiased estimator of the divergence~\cite{FM:guide_and_code}, which is added to the anomaly score to form the likelihood. 
By increasing the number of random vectors $M > 0$, we get a more accurate estimation.

While Hutchinson's estimator draws its $z$ vectors independent of each other, we may try to decrease the variance in our prediction by sampling them from a dependent distribution. 
This is the basis for randomized quasi monte-carlo (RQMC)~\cite{rqmc}.
For the vectors for RQMC we use Latin hypercube sampling~\cite{latin_hypercube}, due to the high dimensionality of our feature vectors.
The results for all estimation methods are shown in~\cref{tab:results_ablation_div_term} for different values for $M$.

Including the divergence does not lead to any noticeable gains for the detection AUROC.
The same behavior applies to all metrics, as completely shown in the supplementary materials.
Most interestingly, with $M = 1$, the divergence only  makes up a total of $.57\%$ of the final score magnitude on Real-IAD, since $\frac{\left|\text{div}\left(u_t^{\theta}\right)\right|}{\left|\log\left( x_0\right)\right|} = 0.0057$.
Moreover, leaving it out gets a boost of at least three times more model executions per second.
Also, increasing $M$ linearly increases both execution time and memory requirements, while also not gaining any precision in detection.
We hypothesize that our model learns a very smooth vector field flowing from $p_0$ to $p_1$, which has almost no noticeable local divergence.
Furthermore, this strengthens us in our decision to leave RealNVP-style models behind, which rely on architectures specifically designed to make this divergence tractable.
Lastly, the flexibility gained with Flow Matching (as was visualized in~\cref{fig:teaser_fm_vs_realnvp}) also provides an advantage on important real-world benchmarks, when comparing it to its more constrained RealNVP competitors.

\subsubsection{Impact of ODE Solver.}\label{sec:ablation_ode}

We test a variety of popular methods for solving ODEs on the Real-IAD data set.
For this we choose methods with fixed step sizes, namely the Euler method, the Midpoint method, and the fourth order variant of Runge-Kutta (RK4)~\cite{ode_hairer}.
We vary their step sizes from $0.2$ to $0.05$ and record their impact on AD performance and model throughput (as FPS) in~\cref{tab:ablation_ode}.
We also compare to the adaptive step version of Runge-Kutta (Dopri5).
Additional adaptive solvers, such as Euler-Heun, Fehlberg2, or Bosh3 achieved similar scores, but are reported in the supplementary for brevity.

Going from Euler to Midpoint to RK4, the amount of function evaluations per batch doubles.
This is also visible in the throughput, which gives the Euler method a clear edge for fast in-production use, since it is the only method capable of processing more than $10$ images per second.
The adaptive Dopri5-solver is slower than all competing methods while not bringing any more precision to the detection.
However, none of the methods show an all-too-clear advantage in the measured AUROC except for the Euler method.
Here, a small improvement can be measured only for the higher step sizes.
All other methods achieve an AUROC of roughly $91.06$, irrespective of step size.

Clear hypotheses can, however, not be made, as the AUROC improvement with Euler lies roughly within the standard deviation reported in~\cref{tab:mean_results_realiad}.
Nevertheless, the simplicity still makes the Euler method the fastest and best usable ODE solver for our AD use case, as there is no trade-off between speed and accuracy.
It is also consistent with previous research, which showed no substantial numerical errors, when using simpler ODE solvers~\cite{FM:seminal}.

\begin{table}[t]
    \begin{minipage}{.49\linewidth}
    \vbox to 6\baselineskip{
        \caption{I-AUROC for the divergence term from~\cref{eq:log_likelihood_equation} on Real-IAD, with varying number of samples $M$. Without the divergence, FPS \& average memory improves, while detection is unchanged.}
        \label{tab:results_ablation_div_term}
        \vfill
      }
      
        \resizebox{\linewidth}{!}{
        \begin{tabular}{cl||c|ccc}
            & Method             & $M=0$ & $M=1$ & $M=2$ & $M=4$ \\\midrule
            \multirow{3}{*}{\rotatebox[origin=c]{90}{\footnotesize{\emph{I-AUROC}}}} & \method        & 91.172  & -       & -       & -       \\[2.25pt]
            & Hutchinson~\cite{hutchinson}        & -              & 91.170  & 91.169  & 91.170  \\[2.25pt]
            & RQMC~\cite{rqmc}                    & -              & 91.170  & 91.170  & 91.170  \\[2.25pt]\midrule
            &  ($\uparrow$) \emph{FPS}            & \textbf{18.77} & 5.40    & 3.16    & 1.56   \\
            & ($\downarrow$) \emph{Memory} (GB)   & \textbf{8.88} & 10.28   & 10.70   & 11.45
        \end{tabular}
        }
        
    \end{minipage}%
    \begin{minipage}{0.01\linewidth}
        \hfill
    \end{minipage}
    \begin{minipage}{.49\linewidth}
      \centering
        \vbox to 6\baselineskip{
        \caption{I-AUROC for different ODE solvers with varying step size parameter on Real-IAD. Increasing the step size also linearly decreases the amount samples processed per second.}
        \label{tab:ablation_ode}
        \vfill
        }
        
        \resizebox{\linewidth}{!}{
        \begin{tabular}{l||ccc|ccc}
            Metric & \multicolumn{3}{c|}{I-AUROC ($\uparrow$)} & \multicolumn{3}{c}{FPS ($\uparrow$)} \\\toprule
            Step size& 0.05  & 0.1   & 0.2   & 0.05 & 0.1  & 0.2   \\ \midrule
            Euler    & 91.09 & \underline{91.12} & \textbf{91.17} & 3.26 & \underline{6.54} & \textbf{13.13} \\
            Midpoint & 91.06 & 91.06 & 91.08 & 1.63 & 3.27 & 6.55  \\
            RK4      & 91.06 & 91.07 & 91.07 & 0.62 & 1.24 & 2.48  \\
            Dopri5   & \multicolumn{3}{c|}{\leftarrowfill 91.06 \rightarrowfill}       & \multicolumn{3}{c}{\leftarrowfill  0.48 \rightarrowfill}     
        \end{tabular}
        }
    \end{minipage} 
\end{table}

\subsubsection{Supplementary Experiments.}
We conduct three additional experiments in our \emph{supplementary} materials in~\cref{sec:appendix_mvtecad,sec:appendix_multiclass,sec:appendix_dinov3}, to further validate our model's strength. 
Firstly, we report metrics for the classic MVTec AD benchmark in the appendix, where \method{} is still competitive.
Furthermore, we also test the popular multi-class AD setting.
Even though \method{} is not explicitly focused on multi-class AD, it achieves a very competitive I-AUROC of $87.5$ on Real-IAD, only slightly beaten by Dinomaly~\cite{dinomaly}.
Lastly, we highlight the robustness to other backbones, by swapping \method{} and PaDiM \& RD4AD to a DINOv3~\cite{dinov3} backbone.
Even then, the AD performance stays the same or deteriorates slightly.
Furthermore, with all methods fixed to a DINOv3 backbone, \method{} still outperforms the other methods.
Still, we prefer the classic WideResNet due to its lower computational costs.


%% file: sec/5_conclusion.tex
\section{Conclusion}\label{sec:conclusion}

We have proposed~\method, an effective framework for detecting anomalies in the multi-view setting.
Using the simple Flow Matching formulation and a multi-layer feature extraction pipeline, we are able to establish state-of-the-art results on Real-IAD, as well as new comprehensive benchmarking results on MANTA-Tiny.
Future work will focus on including more prior knowledge, such as explicit 3D constraints or multi-view correspondences, into the Flow Matching pipeline.
More recent improvements in Flow Matching and generative modelling, such as energy matching~\cite{FM:energy_matching}, may also suit themselves to future endeavours in anomaly detection.  


%% file: sec/X_suppl.tex
\clearpage
\setcounter{page}{1}
\setcounter{section}{0}
\renewcommand{\thesection}{\Alph{section}}

\renewcommand{\theHsection}{supp.\thesection}
\renewcommand{\theHfigure}{supp.\arabic{figure}}
\renewcommand{\theHtable}{supp.\arabic{table}}

\title{Supplementary: {MATCH}: Flow Matching for Multi-View Anomaly Detection} 
\author{}\authorrunning{}\institute{}\email{}
\titlerunning{Supplementary: {MATCH}: Flow Matching for Multi-View AD}
\maketitle

\section{Architecture Details}\label{sec:appendix_arch}

We describe the architecture in greater detail, noting down all layers and their configurations.
The respective input dimension, output dimension, and stride (\eg s1 for a stride of $1$) are shown below the convolutional blocks.
"Conv" describes regular, while "ConvT" are transposed convolutions.
The Group Normalization ("GN") blocks also show the number of groups (always 32), as well as the feature dimension.
Occasionally, tensor dimensions are displayed below the data nodes.

\begin{wrapfigure}{r}{0.5\linewidth}
    \vspace{-2\baselineskip}
    \centering
    \includegraphics[width=\linewidth]{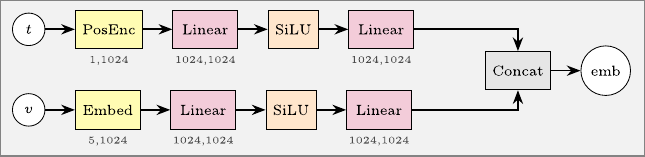}
    \caption{Embedding Network, with positional encoding (PosEnc) and embedding layers (Emb).}
    \label{fig:appendix_embedding}
    
    \vspace{-\baselineskip}
\end{wrapfigure}

\paragraph{Embeddings.}
The short MLPs for embedding time and view information are shown in~\cref{fig:appendix_embedding}.

\begin{figure}[t]
    \centering
    \includegraphics[width=0.95\linewidth]{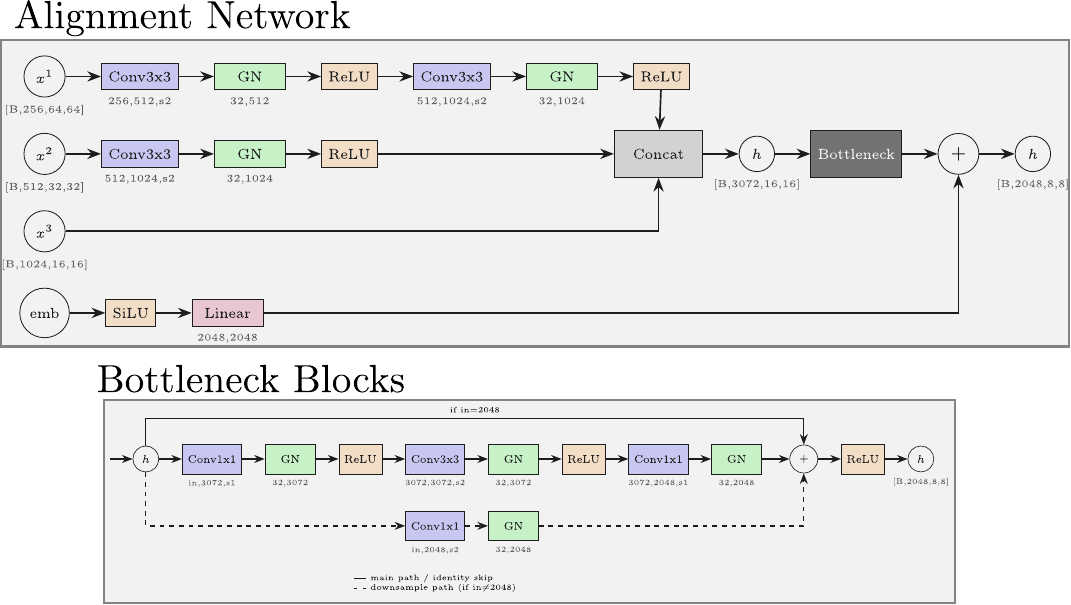}
    \caption{Alignment and Bottleneck Network}
    \label{fig:appendix_bottleneck}
\end{figure}

\paragraph{Bottleneck.}
The full bottleneck, including feature alignment and bottleneck blocks, is shown in~\cref{fig:appendix_bottleneck}.
Three feature vectors $x^i$ are aligned by convolutional layers and then concatenated.
After adding the already concatenated embedding of view and time information ("emb"), they are passed to \emph{three} bottleneck blocks, with input dimensions $3072$, $2048$, and $2048$ respectively.
This results in the first block needing to use downsampling for its residual connection, while the others use identity residual connections.

\begin{figure}[t]
    \centering
    \includegraphics[width=0.8\linewidth]{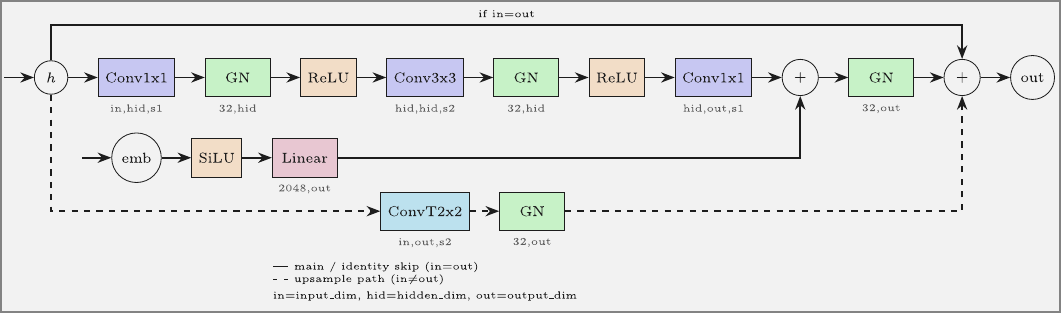}
    \caption{ResNet-style Decoder Block, depending on choice of input (in), hidden (hid) and output (out) dimension.}
    \label{fig:appendix_decoder}
\end{figure}

\paragraph{Decoder.} 
A decoder block is shown in~\cref{fig:appendix_decoder}.
Again, whenever input and output dimension do not match (which is the case for every first block), the residual connection is done with a convolutional upsampling.
Each first block (with the downsampling residual connection), also uses ConvT2x2 instead of Conv3x3. 
The output of one block serves as input for the next.

Each decoder block is determined by a $(\text{input}, \text{hidden}, \text{output})$-tuple of dimensions.
Additionally, the embedding is also added onto the resulting feature.
The mentioned blocks $D_i$ of~\cref{fig:rd_architecture} consist of three, four, and six blocks respectively.
Thus $D_1$ has a $(2048, 3072, 1024)$ block, followed by two $(1024,3072,1024)$ blocks.
$D_2$ contains a $(1024,1536,512)$ block and then three $(512,1536,512)$ blocks.
Lastly, $D_3$ consists of a $(512,768,256)$ block and five $(256,768,256)$ blocks.

Changing the width\_per\_group parameter, currently set to $768$ and ablated in~\cref{fig:ablation_dimensionality} in the main paper, controls the width of these layers.  
Each decoder $D_i$ produces an output $\hat{x}^i$ of the same shape as the respective input feature $x^i$, with which the Flow Matching loss is calculated.

\section{Additional Multi-Class Anomaly Detection}\label{sec:appendix_multiclass}

Additional evaluation is done with the popular multi-class AD setting, despite \method{} not being explicitly designed for this task.
Here, all training splits of the classes of Real-IAD are combined. 
After training for $100$ epochs, all classes are evaluated independently.

The average performance across all classes is shown in~\cref{tab:appendix_multiclass_ad_realiad}.
Baseline numbers are taken from multi-class AD method Dinomaly~\cite{dinomaly}, which takes some numbers from the ADer benchmark paper~\cite{ader}.
We outperform all regular single-class methods by large margins, while being competitive with the multi-class baselines.
Concretely, we achieve an I-AUROC of $87.5$, only lacking two points behind Dinomaly.

\begin{table}[t]
  \centering
  \caption{Performance under \textbf{multi-class} AD setting for Real-IAD. Numbers are taken from the Dinomaly paper~\cite{dinomaly}. Methods designed for the multi-class setting are marked in grey. The best performing method is marked in \textbf{bold} with the runner-up \underline{underlined}.}
    \begin{tabular}{l||c|c|c}
    \multicolumn{1}{c||}{} & \multicolumn{3}{c}{Metrics ($\uparrow$)} \\\toprule
    Method      & I-AUROC           & P-AUROC           & P-AUPRO           \\\midrule     
     RD4AD~\cite{rd}                      &82.4  & 97.3  & 89.6 \\
     SimpleNet~\cite{simplenet}        &57.2  & 75.7  & 39.0 \\
     DeSTSeg~\cite{destseg}            &82.3  & 94.6  & 40.6 \\\hline
     \rowcolor{gray!25}UniAD~\cite{uniad}            &83.0  & 97.3  & 86.7\\
     \rowcolor{gray!25}ReContrast~\cite{reconstrast} &86.4  & 97.8  & 91.8 \\
     \rowcolor{gray!25}DiAD~\cite{diAD}              &75.6  & 88.0  & 58.1 \\
     \rowcolor{gray!25}ViTAD~\cite{zhang2024}        &82.7  & 97.2  & 84.8 \\
     \rowcolor{gray!25}MambaAD~\cite{mambaad}  &86.3  & \underline{98.5}  & 90.5 \\
     \rowcolor{gray!25}MVAD~\cite{MVAD} & 86.6 & 97.9 & 91.2 \\
     \rowcolor{gray!25}Dinomaly~\cite{dinomaly}      &\textbf{89.3}  & \textbf{98.8}  & \textbf{93.9}  \\\hline
     \method{} (ours)                  &\underline{87.5}  & 98.0  & \underline{92.2} \\
    \end{tabular}
  \label{tab:appendix_multiclass_ad_realiad}
\end{table}

\section{Additional Results on MVTec AD}\label{sec:appendix_mvtecad}
\begin{wrapfigure}{r}{0.3\linewidth}
    \vspace{-2\baselineskip}
    \centering
    \includegraphics[width=\linewidth]{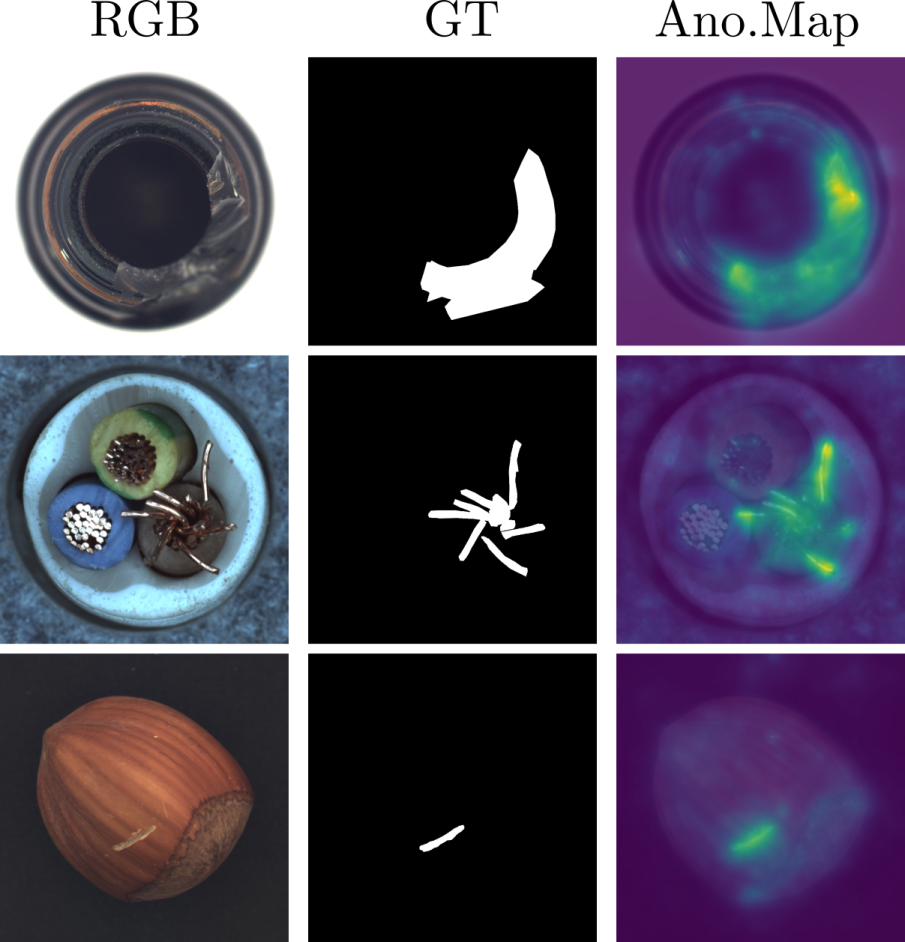}
    \caption{Qualitative results for the MVTec AD data set~\cite{dataset:mvtec} on the classes \emph{bottle}, \emph{cable}, and \emph{hazelnut}.~\method{} consistently segments the anomalous regions.}
    \label{fig:appendix_mvtec_qualitative}
    \vspace{-2\baselineskip}
\end{wrapfigure}

We report results on the popular MVTec AD data set~\cite{dataset:mvtec} in~\cref{tab:mvtec2d_result}.
Since we did not optimize our method specifically for this data set, we do not reach peak state-of-the-art performance, but are still comfortably competitive, with an I-AUROC of $97.46$ and a P-AUPRO of $93.88$. 
Some qualitative examples are found in~\cref{fig:appendix_mvtec_qualitative}.

\section{Additional Results for DINOv3 Backbone}\label{sec:appendix_dinov3}

We also test the new DINOv3~\cite{dinov3} model as a backbone for \method as well as baselines RD4AD~\cite{rd} and PaDiM~\cite{padim}.
We chose these baselines, since they most naturally support swapping to a different custom backbone.
Since we still restrict ourselves to consumer-level hardware, we use the DINOv3 ViT-S/16 model, which is pretrained on the enormous LVD-1689M data set.
The results are presented in~\cref{tab:appendix_dinov3}.
As \method{} still performs strongest, we confirm its robustness to different and newer backbones.
However, we still prefer to use our WideResNet, since it achieves similar or stronger scores, while being less computationally expensive.
Expanding to different backbones or even learning on raw pixel scores could therefore be interesting avenues for future research.
Further, introducing huge pretraining data sets may not necessarily lead to an increase in performance. 

\begin{table}[t]
\centering
\caption{Detection and segmentation metrics on MVTec AD~\cite{dataset:mvtec}.}
\begin{tabular}{l||ccc}
\multicolumn{1}{c||}{} & \multicolumn{3}{c}{Metrics ($\uparrow$)} \\\toprule
Category   & I-AUROC & P-AUROC & P-AUPRO  \\\midrule
carpet     & 98.84   & 98.96   & 96.31    \\
grid       & 99.00   & 99.15   & 97.02    \\
leather    & 99.97   & 99.23   & 98.03    \\
tile       & 99.78   & 93.41   & 93.76    \\
wood       & 99.39   & 94.09   & 89.61    \\\midrule
bottle     & 99.84   & 98.14   & 95.64    \\
cable      & 92.95   & 96.23   & 89.65    \\
capsule    & 97.21   & 99.05   & 94.69    \\
hazelnut   & 99.93   & 98.43   & 93.38    \\
metal nut  & 99.90   & 97.22   & 94.00    \\
pill       & 95.36   & 96.96   & 96.73    \\
screw      & 97.91   & 99.6    & 97.71    \\
toothbrush & 90.28   & 98.82   & 95.41    \\
transistor & 97.04   & 90.37   & 81.11    \\
zipper     & 94.54   & 98.53   & 95.21    \\\midrule
MVTec AD   & 97.46   & 97.21   & 93.88   
\end{tabular}
\label{tab:mvtec2d_result}
\end{table}

\begin{table}[t]
    \centering
    \caption{Anomaly Detection on Real-IAD with \textbf{DINOv3}. We extract features from the layers [1, 6, 11].}
    \begin{tabular}{c|c||cccccc}
         Backbone & Model & S-AUROC        & I-AUROC        & P-AP           & P-AUPRO        & P-AUROC         \\\midrule
         \multirow{3}{*}{\textbf{DINOv3}} & PaDiM & \uline{93.64}  & 79.92          & 07.20            & 75.88          & 94.83           \\
          & RD4AD & 92.93          & \uline{86.58}  & \uline{21.73}  & \uline{83.49}  & \uline{97.15}   \\
          & \method & \textbf{95.58} & \textbf{88.63} & \textbf{26.08} & \textbf{87.83} & \textbf{98.38}  \\\midrule
         \rowcolor{gray!25} W.ResNet &  \method & 95.63 & 91.17 & 32.68 & 94.76 & 99.24
\end{tabular}    
    \label{tab:appendix_dinov3}
\end{table}

\section{Two Moons Toy Experiment}\label{sec:appendix_twomoons}

We describe the setting for the toy experiment with the "Two Moons" data set shown in~\cref{fig:teaser_fm_vs_realnvp} (main paper).
While the data itself is two-dimensional, we linearly embed it in $d$-dimensional space by sampling a random matrix in $\mathbb{R}^{d\times d}$, orthogonalizing it with QR decomposition and using the first $n = 2$ dimensions for embedding our data.
We use our flow models to map the moons, lifted to $d$ dimensions, onto a $d$-dimensional multivariate standard Gaussian, from which we will later sample new instances.
For visualization, we use the inverse of our embedding to go back to two dimensions.
For our experiments, we use the values $d \in \{2, 8, 32, 128,256,512\}$.
Both models use the Adam~\cite{adam} optimizer with a learning rate of $0.001$ and $\beta_{1,2} = \left(0.9, 0.999\right)$.
We train for $50$k epochs with a batch size of $512$.

For \emph{Flow Matching}, we use the same loss as in~\cref{sec:method}.
The model is a series of six simple four-layer MLPs, with a hidden dimension of $32$ and SELU activation, which has them ranging between 20k and 39k parameters, depending on $d$.

The \emph{RealNVP}~\cite{realnvp} model consists of $8$ coupling blocks trained with a maximum likelihood loss.
These blocks have a fixed random permutation, which simulates random masking needed to have all features interact with each other. 
Each coupling block has a network for scaling $s$ and translation $t$. These are implemented as four-layer MLPs with LeakyReLU activation, layer normalization and the same hidden dimension of $32$.
Clamping the scale factors $s$ to the interval $[-3,3]$, as suggested by Ardizzone et al.~\cite{ardizzone_nf}, further increases training stability.
Since RealNVP needs to preserve the dimension of the input after every coupling block, the models lie in the range of 38k to roughly 170k parameters, which is much larger than their respective Flow Matching counterparts.

\subsection{Quantitative Experiment}

We quantitatively measure the quality of our density estimation using bits per dimension (BPD)~\cite{glow_openai}, which is the normalized average negative log likelihood.

For the Two Moons experiment, the FM model achieves consistently sensible BPD values (in the range of $1.3$ to $2.0$) for all dimensions.
RealNVP, on the other hand, produces degenerate (negative) BPD values in all dimensions $> 2$.

Additionally, we want to evaluate the quality of the density estimation on the actual feature space used by~\method{}.
For this, we construct a RealNVP-based normalizing flow baseline, which is based on the strong teacher network of~\cite{AST_rudolph}.
We use one teacher network for each of the feature-map scales, with four coupling blocks and a hidden dimension of $256$ each.
We follow all settings of Rudolph~\etal~\cite{AST_rudolph}, aggregating features across scales using the maximum operator.
Likelihoods for~\method{} are calculated using Hutchinson's trace estimator (with $M=1$ sample).

\cref{tab:bpd_wideresnet}~ shows the result for both AD and density estimation.
The NF baseline achieves non-trivial I-AUROC (even surpassing some other baselines), while its BPD of $160.5$ is vastly higher than MATCH, which lies at $1.83$.
Additionally, Multi-Flow, which is trained on a different feature-space, also reaches a higher BPD of $2.10$.
This confirms the FM formulation as better suited for not only the AD task, but also more capable for density estimation in these feature spaces.

\begin{table}[t]
    \centering
    \caption{Comparing RealNVP and Flow Matching on Real-IAD for AD \& density estimation. Only Multi-Flow operates in a \textcolor{gray}{different feature space}.}
    \label{tab:bpd_wideresnet}
    \begin{tabular}{c|l|c|c}
        Objective&Method & I-AUROC ($\uparrow$) & BPD ($\downarrow$) \\\midrule
         \multirow{2}{*}{RealNVP}& \cellcolor{gray!25} Multi-Flow & \cellcolor{gray!25} 90.27  &  \cellcolor{gray!25} 2.10   \\
         &Baseline (NF) & 80.91 & 160.5 \\\hline 
        FM&\method & \textbf{91.17}  & \textbf{1.83} \\
    \end{tabular}
\end{table}

\section{Additional Qualitative Results}\label{sec:appendix_qualitative}

\begin{figure}
    \centering
    \includegraphics[width=\linewidth]{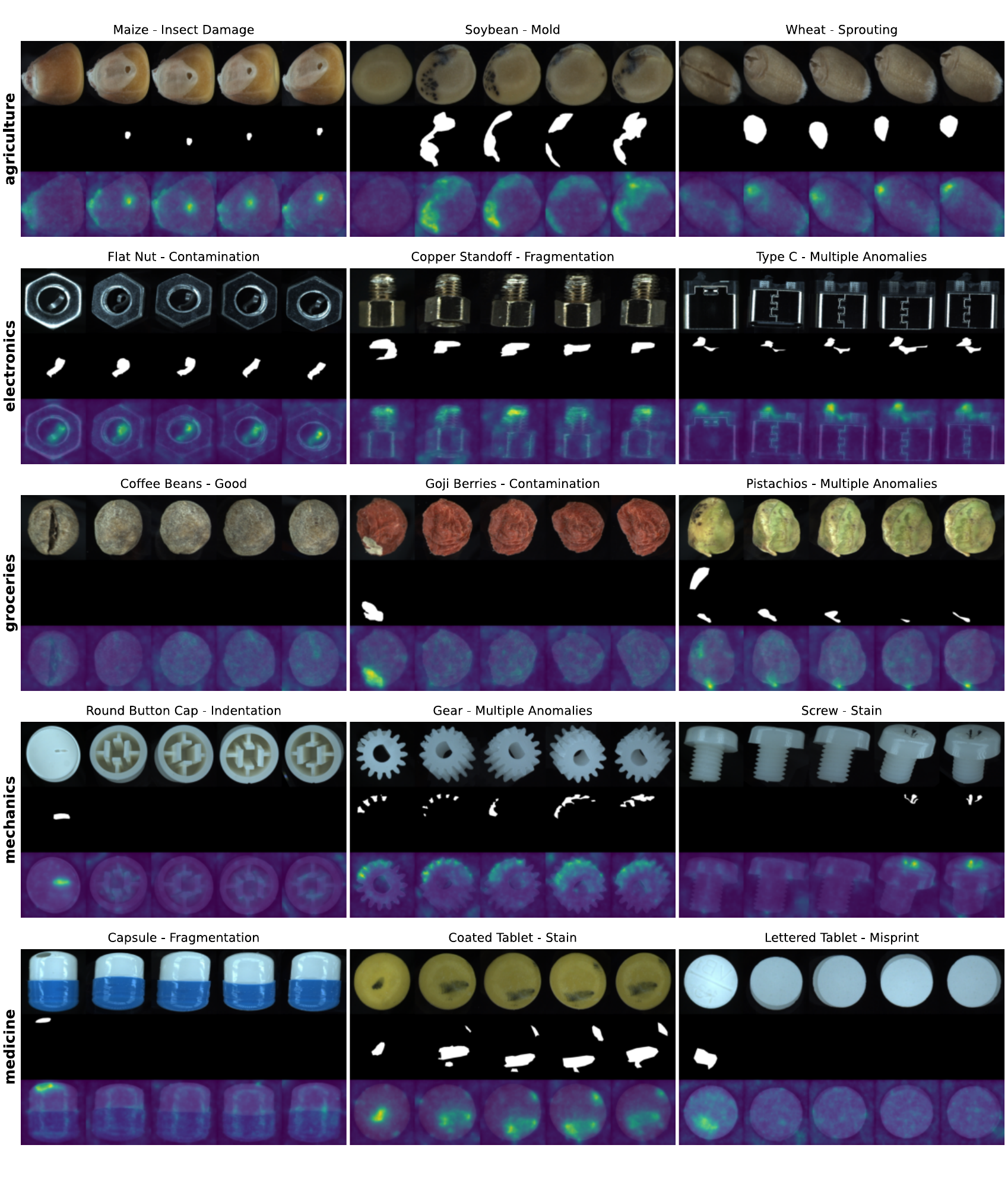}
    \caption{Qualitative anomaly maps for every category of MANTA-Tiny, with class and anomaly type designated. Despite the vast variety of products, \method{} regularly segments the anomalies, while avoiding excessive false positives.}
    \label{fig:appendix_qualitative_mantatiny}
\end{figure}

We provide a larger qualitative overview of well-segmented anomalies in~\cref{fig:appendix_qualitative_mantatiny}.

\paragraph{Failure Cases.}

\begin{figure}[htbp]
    \centering
    \begin{subfigure}[b]{0.32\textwidth}
        \centering
        \includegraphics[width=\textwidth]{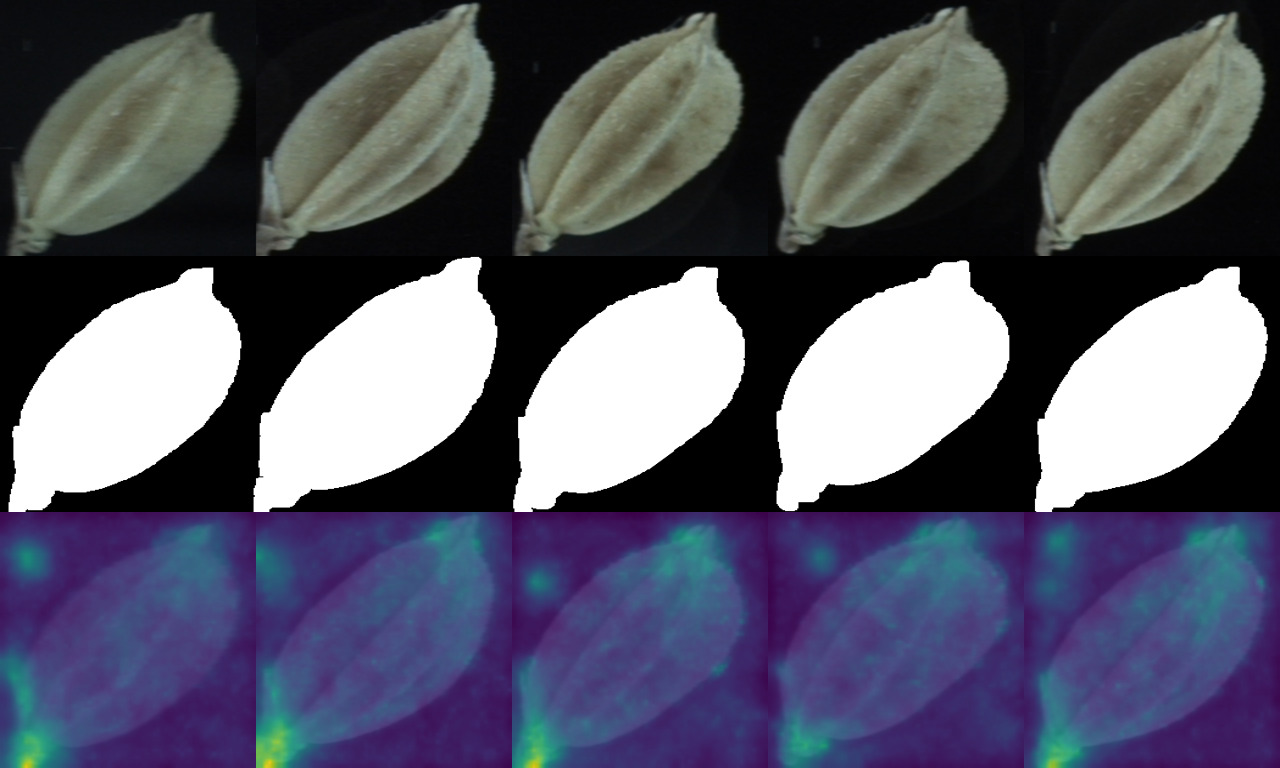}
        \caption{Paddy - Shriveling}
        \label{fig:failure_cases_subfig1}
    \end{subfigure}
    \hfill
    \begin{subfigure}[b]{0.32\textwidth}
        \centering
        \includegraphics[width=\textwidth]{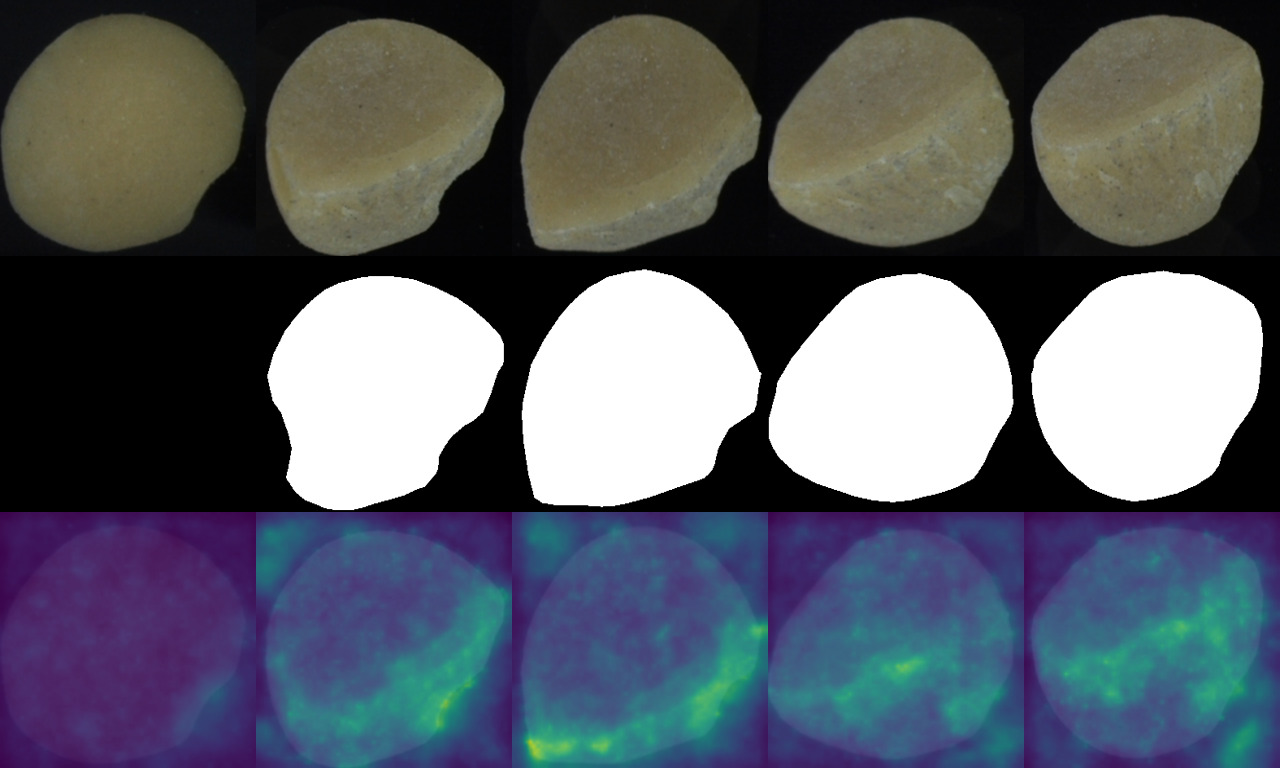}
        \caption{Soybean - Crack}
        \label{fig:failure_cases_subfig2}
    \end{subfigure}
    \hfill
    \begin{subfigure}[b]{0.32\textwidth}
        \centering
        \includegraphics[width=\textwidth]{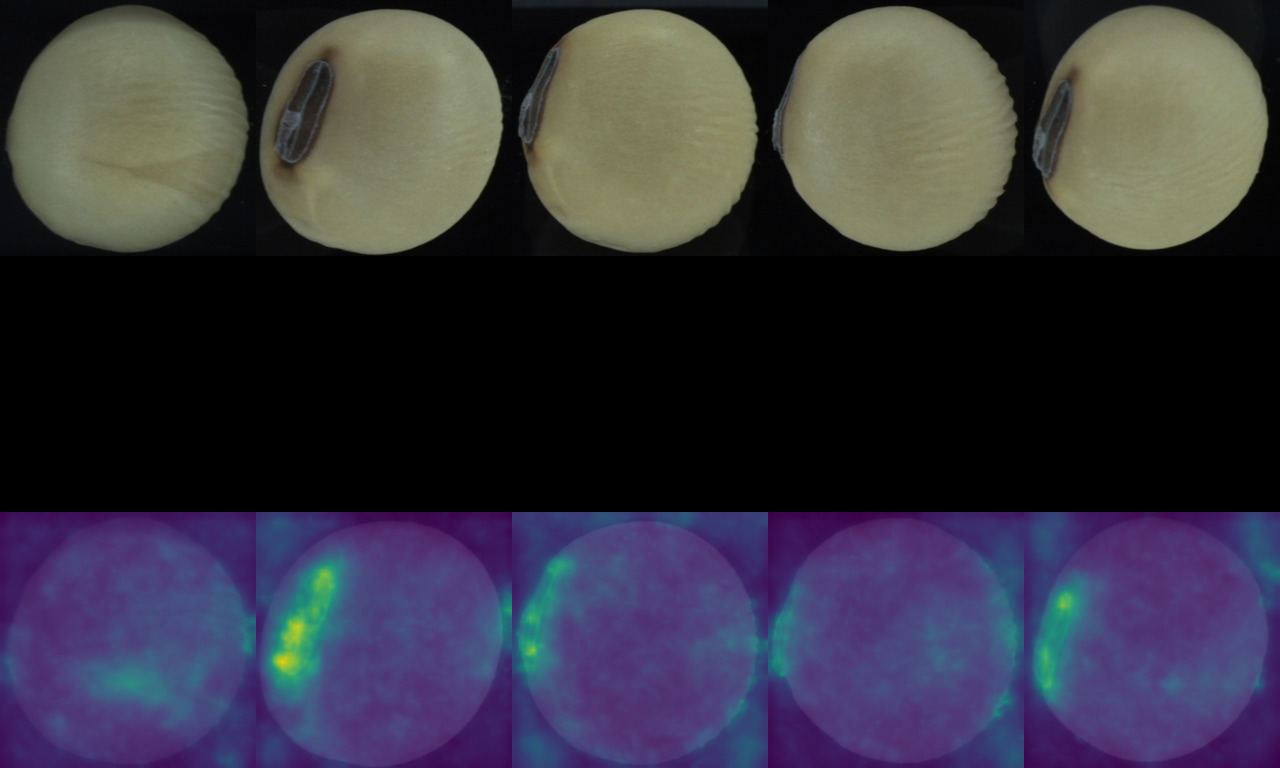}
        \caption{Soybean - Good}
        \label{fig:failure_cases_subfig3}
    \end{subfigure}
    
    \vspace{0.5em}

    \begin{subfigure}[b]{0.32\textwidth}
        \centering
        \includegraphics[width=\textwidth]{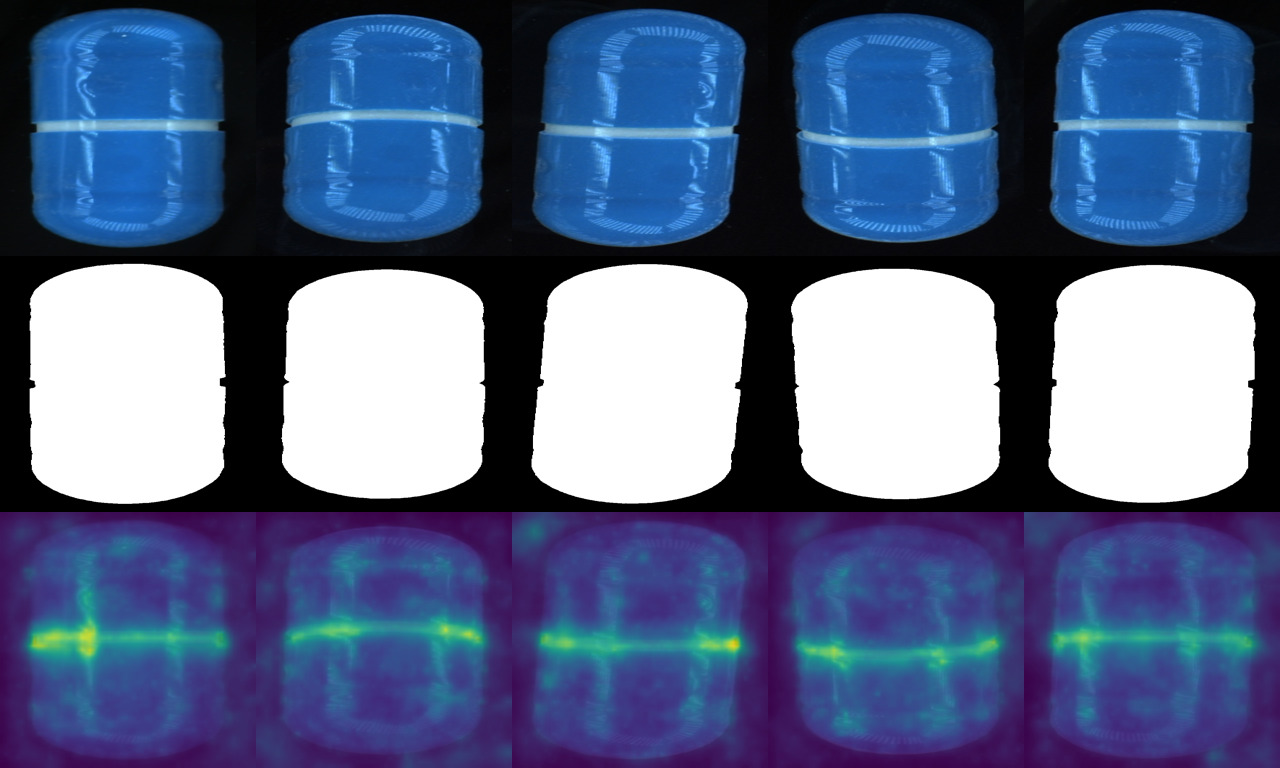}
        \caption{Capsule - Double Cap}
        \label{fig:failure_cases_subfig4}
    \end{subfigure}
    \hfill
    \begin{subfigure}[b]{0.32\textwidth}
        \centering
        \includegraphics[width=\textwidth]{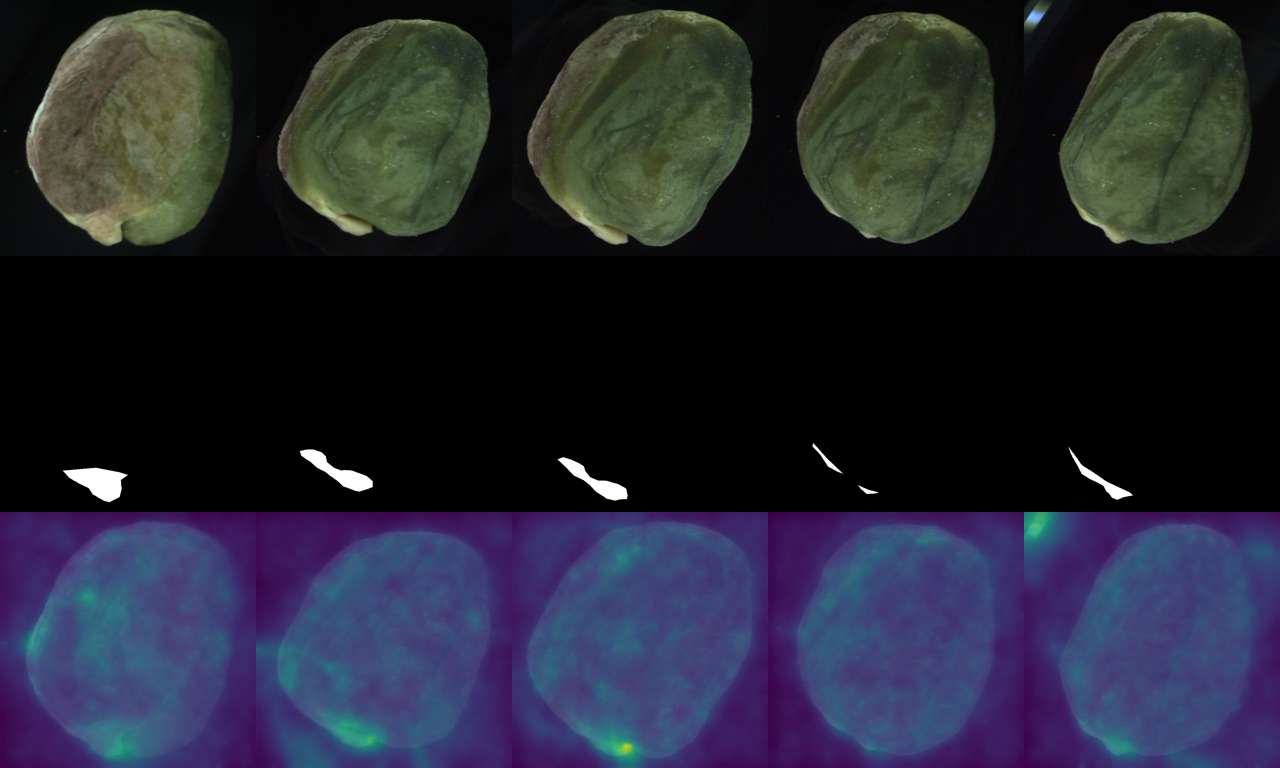}
        \caption{Pistachios - Sprouting}
        \label{fig:failure_cases_subfig5}
    \end{subfigure}
    \hfill
    \begin{subfigure}[b]{0.32\textwidth}
        \centering
        \includegraphics[width=\textwidth]{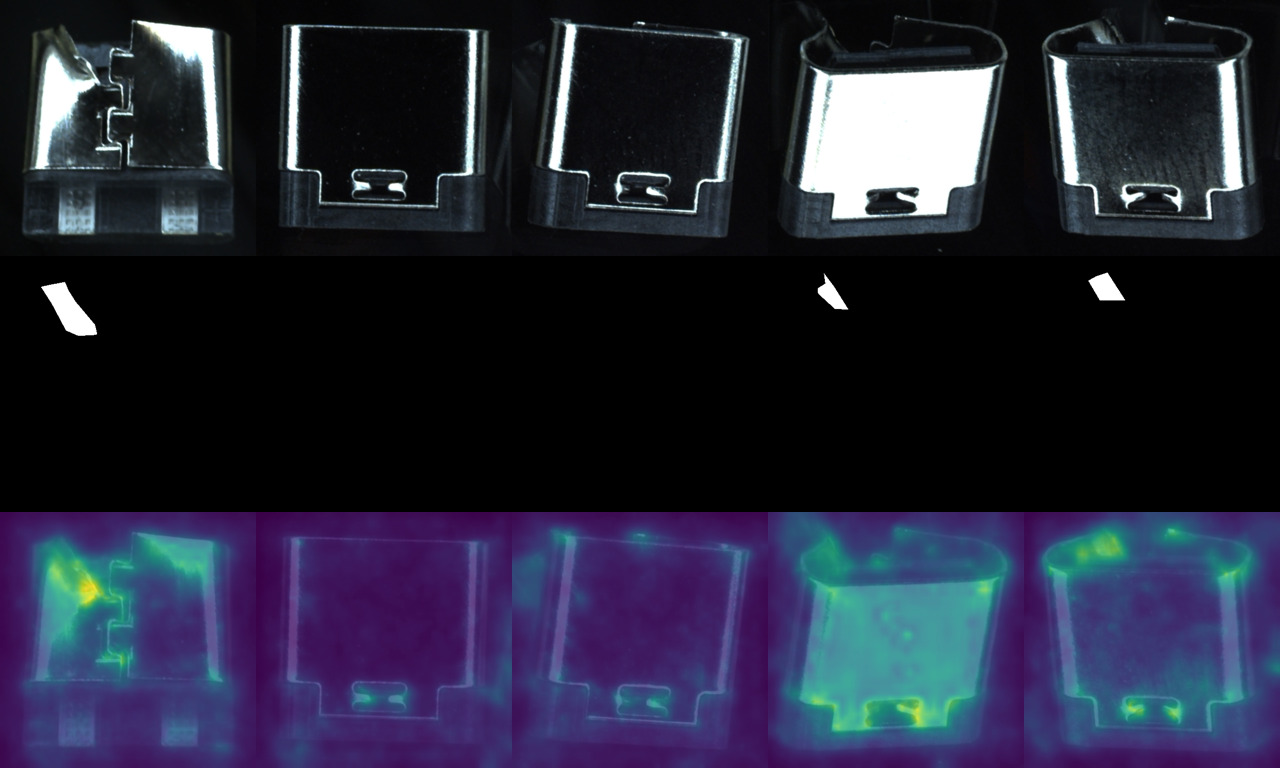}
        \caption{Type C - Protrusion}
        \label{fig:failure_cases_subfig6}
    \end{subfigure}
    
    \caption{Anomaly maps of the most common failure cases in MANTA-Tiny.}
    \label{fig:appendix_failure_cases}
\end{figure}

We present a variety of common failure cases of \method{} in \cref{fig:appendix_failure_cases}, discuss them here and point towards possible solutions for future models.

Both \cref{fig:failure_cases_subfig1,fig:failure_cases_subfig2} show  very large ground truth anomalies, which are not fully segmented. 
The "Shriveling" anomaly is caused by the raw paddy rice being less wet and underdeveloped.
However, the overall structure remains intact, while only a very slight deviance in colour needs to be detected. 
Due to the very large and diverse training set, detecting this anomaly is magnitudes harder than most of the industrial anomalies, which have less ambiguity in colour.

As for the crack in the soybean in~\cref{fig:failure_cases_subfig2}, the crack itself is properly segmented. 
However, with large chunks missing, the entire soybean is now considered an anomaly, even when locally it may appear intact.

\Cref{fig:failure_cases_subfig3} presents a false positive, which consists of a black spot (where the soybean was once connected to the plant).
This black spot is also visible in some of the training images and is naturally not considered an anomaly.
However, since this spot is rather small, most of the training data does not show it, making it rarer and therefore naturally more prone to be close to the decision boundary and a false positive.
Here, data augmentation or focussing more on sub-optimally covered areas in the training set may help in eliminating this error.

\Cref{fig:failure_cases_subfig4} shows a capsule with two blue caps, whereas one cap should be white.
Again, the entire object is considered an anomaly.
Each of the blue caps could, by themselves, be considered locally plausible.
Only in their global combination do they appear as an anomaly.
However, the segmentation is currently more focused towards local defects, while detection is reasoning more globally.
Informing the segmentation with the more global anomaly score could improve the model here.

\Cref{fig:failure_cases_subfig5,fig:failure_cases_subfig6} both present false positives caused by the ambiguity of the data.
The last view of~\cref{fig:failure_cases_subfig5} has a small foreign object in the corner, that is not part of the anomaly mask.
This failure may be alleviated by segmenting the objects before processing them.
In~\cref{fig:failure_cases_subfig6}, the object is rotated in a way such that it reflects the light from source into the camera.
This lighting is never found in the training data and thereby it is treated as an anomaly by the model, even when the 3D structure is plausible.
Additional 3D scans/depth estimation could help circumventing the issue of lighting irregularities.

\newpage

\section{Comprehensive Experimental Results}\label{sec:appendix_comprehensive}

\paragraph{Comprehensive Ablations.}
The impact of removing the divergence on every metric (as described in~\cref{sec:ablation_divergence} in the main paper) is shown in~\cref{tab:appendix_ablation_div}. We executed this experiment with a batch size of $2$, and record the average GPU memory usage across the entire test set. 

All metrics for the ablation of the ODE solvers from~\cref{sec:ablation_ode} are shown in~\cref{tab:appendix_ablation_ode}.

\paragraph{Comprehensive experiments on Real-IAD and MANTA-Tiny.}

The results for the test splits of MANTA versus MANTA-Tiny are shown in~\cref{tab:appendix_manta_full}.

We also provide the comprehensive results of every model tested on every object category/class for the metrics I-AUROC, S-AUROC, P-AUPRO, P-AUROC, and P-AP on Real-IAD and MANTA-Tiny.
Metrics for Real-IAD are reported in~\cref{tab:appendix_realiad_i-auroc,tab:appendix_realiad_s-auroc,tab:appendix_realiad_p_aupro,tab:appendix_realiad_p_auroc,tab:appendix_realiad_p_ap}, while~\cref{tab:appendix_manta_i-auroc,tab:appendix_manta_s-auroc,tab:appendix_manta_p-aupro,tab:appendix_manta_p-auroc,tab:appendix_manta_p-ap} show results on MANTA-Tiny.

\begin{table}[ht]
\centering
\caption{Influence of the divergence calculation method on every metric, including memory and FPS. The divergence has roughly no influence on detection and segmentation metrics.}
\label{tab:appendix_ablation_div}
\begin{tabular}{c|c||c|c|c|c|c|c|c}
        Method    & $M$                & S-AUROC & I-AUROC& P-AP   & P-AUPRO & P-AUROC&      FPS ($\uparrow$) & Mem. ($\downarrow$)     \\\midrule
        Hutch.    & \multirow{2}{*}{1} & 95.614  & 91.170 & 32.657 & 94.749  & 99.243 & \multirow{2}{*}{5.40} & \multirow{2}{*}{10.28 GB}          \\
        RQMC      &                    & 95.614  & 91.170 & 32.658 & 94.749  & 99.243 &                       & \\\midrule
        Hutch.    & \multirow{2}{*}{2} & 95.615  & 91.169 & 32.658 & 94.749  & 99.243 & \multirow{2}{*}{3.16} & \multirow{2}{*}{10.70 GB}           \\
        RQMC      &                    & 95.615  & 91.170 & 32.658 & 94.749  & 99.243 &                       & \\\midrule
        Hutch.    & \multirow{2}{*}{4} & 95.615  & 91.170 & 32.657 & 94.749  & 99.243 & \multirow{2}{*}{1.56} & \multirow{2}{*}{11.45 GB}           \\
        RQMC      &                    & 95.617  & 91.170 & 32.658 & 94.749  & 99.243 &                       & \\\midrule
        \method   & -                  & 95.632  & 91.172 & 32.678 & 94.763  & 99.241 & \textbf{18.77}        & \textbf{8.88 GB} 
\end{tabular}
\end{table}

\begin{table}
\centering
\caption{All metrics for different ODE solvers with varying step size parameter on Real-IAD.}
\label{tab:appendix_ablation_ode}
\begin{tabular}{l||c|c|c|c|c|c}
Solver         & Step size & I-AUROC & P-AP   & P-AUPRO & P-AUROC & FPS    \\\midrule
Euler          & 0.05     & 91.088  & 32.428 & 94.698  & 99.234  & 3.26   \\
Euler          & 0.10     & \uline{91.117} & 32.506 & \uline{94.721}  & \uline{99.237}  & \uline{6.54}   \\
Euler          & 0.20     & \textbf{91.172}  & \textbf{32.678} & \textbf{94.763}  & \textbf{99.241}  & \textbf{18.77}  \\\hline
Midpoint       & 0.05     & 91.059  & 32.355 & 94.680  & 99.231  & 1.63   \\
Midpoint       & 0.10     & 91.059  & 32.338 & 94.679  & 99.231  & 3.27   \\
Midpoint       & 0.20     & 91.076  & 32.328 & 94.690  & 99.230  & 6.55   \\\hline
RK4            & 0.05     & 91.063  & \uline{32.360} & 94.683  & 99.232  & 0.62   \\
RK4            & 0.10     & 91.065  & 32.365 & 94.685  & 99.232  & 1.24   \\
RK4            & 0.20     & 91.066  & 32.367 & 94.687  & 99.231  & 2.48   \\\midrule
Adaptive-Heun  & -        & 91.063  & \uline{32.360} & 94.681  & 99.231  & 0.35   \\
Bosh3          & -        & 91.063  & \uline{32.360} & 94.681  & 99.231  & 0.46   \\
Dopri5         & -        & 91.063  & 32.359 & 94.681  & 99.231  & 0.48  \\
Fehlberg2      & -        & 91.062  & 32.359 & 94.681  & 99.231  & 1.60   
\end{tabular}
\end{table}

\begin{table}
\centering
\caption{Testing \method{} (trained on MANTA-Tiny) on the test splits of both MANTA and MANTA-Tiny. There are no clear tendencies towards a worse performance on the whole MANTA split. Thereby, MANTA-Tiny is a suitable alternative to the entire MANTA split.}
\label{tab:appendix_manta_full}
\begin{tabular}{l||c|c|c|c|c}
Test Split & I-AUROC & S-AUROC & P-AP  & P-AUPRO & P-AUROC  \\\midrule
MANTA-Tiny & 90.66   & 89.41   & 40.48 & 89.66   & 95.65    \\
MANTA      & 90.74   & 89.47   & 42.95 & 89.23   & 95.61   
\end{tabular}
\end{table}

\begin{table*}[ht]
    \centering
    \caption{Comprehensive results for anomaly detection, measured as \textbf{I-AUROC} ($\uparrow$) on Real-IAD. The best performing method is marked in \textbf{bold} with the runner-up \underline{underlined}. Our method is run $n=5$ times.}
    \resizebox{\linewidth}{!}{
    \begin{tabular}{l||c|c|c|c|c|c|c|c|c|c}
    \multirow{2}{*}{Category}    &  CFA   & DSR   & EfficientAD & FastFlow & Multi-Flow & PaDiM & RD4AD    & SimpleNet & STFPM & \method          \\
    & \cite{cfa} & \cite{zavrtanik2022dsr} & \cite{efficientAD} & \cite{fastflow} & \cite{MultiFlow} & \cite{padim} & \cite{rd} & \cite{simplenet} & \cite{stfpm} & \emph{(ours)} \\\midrule
        Audiojack         & 79.25 & 60.92 & 79.81 & 69.94 & \textbf{90.69} & 68.57 & 84.14          & 77.47          & 79.17 & \uline{87.61}  \scriptsize{$\pm$ 0.13}  \\
        Bottle Cap        & 86.96 & 53.80 & 80.01 & 86.26 & 94.62          & 82.51 & \uline{95.43}  & 91.46          & 63.53 & \textbf{96.41} \scriptsize{$\pm$ 0.03}  \\
        Button Battery    & 77.90 & 77.60 & 83.24 & 85.01 & 89.37          & 85.53 & \uline{91.22}  & 91.06          & 74.59 & \textbf{91.92} \scriptsize{$\pm$ 0.07}  \\
        End Cap           & 71.78 & 53.25 & 81.47 & 69.88 & 84.77          & 75.93 & \uline{85.95}  & 53.14          & 60.25 & \textbf{87.31} \scriptsize{$\pm$ 0.11}  \\
        Eraser            & 78.29 & 47.99 & 77.04 & 79.84 & 92.08          & 84.13 & \uline{92.28}  & 91.06          & 86.66 & \textbf{94.19} \scriptsize{$\pm$ 0.08}\\
        Fire Hood         & 73.86 & 57.24 & 69.48 & 75.63 & \textbf{93.35} & 72.75 & 82.88          & 81.56          & 77.23 & \uline{87.22}  \scriptsize{$\pm$ 0.09}  \\
\end{tabular}
    }
    \label{tab:appendix_realiad_i-auroc}
\end{table*}

\begin{table*}[ht]
    \centering
    \caption{Comprehensive results for anomaly detection, measured as \textbf{S-AUROC} ($\uparrow$) on Real-IAD. The best performing method is marked in \textbf{bold} with the runner-up \underline{underlined}. Our method is run $n=5$ times.}
    \resizebox{\linewidth}{!}{
    \begin{tabular}{l||c|c|c|c|c|c|c|c|c|c}
    \multirow{2}{*}{Category}    &  CFA   & DSR   & EfficientAD & FastFlow & Multi-Flow & PaDiM & RD4AD    & SimpleNet & STFPM & \method          \\
    & \cite{cfa} & \cite{zavrtanik2022dsr} & \cite{efficientAD} & \cite{fastflow} & \cite{MultiFlow} & \cite{padim} & \cite{rd} & \cite{simplenet} & \cite{stfpm} & \emph{(ours)} \\\midrule
        Audiojack         & 74.81 & 80.11 & 89.91          & 93.14          & \textbf{96.56} & 92.96          & \uline{94.77}  & 91.20           & 43.73 & 90.66           \scriptsize{$\pm$0.20}   \\
        Bottle Cap        & 97.07 & 61.69 & \textbf{99.45} & 97.08          & 98.00          & 97.50          & 98.23          & \uline{99.40}   & 61.45 & 98.50           \scriptsize{$\pm$0.06}  \\
        Button Battery    & 92.91 & 66.70 & 90.58          & \uline{96.40}  & \textbf{96.57} & 89.35          & 95.39          & 95.80           & 69.83 & 96.04           \scriptsize{$\pm$0.23}  \\
        End Cap           & 77.26 & 39.08 & 88.65          & 85.60          & 90.89          & 74.01          & 93.24          & \uline{94.20}   & 52.17 & \textbf{94.47}  \scriptsize{$\pm$0.13}  \\
        Eraser            & 72.52 & 30.07 & 92.69          & 92.31          & 92.81          & \textbf{97.22} & 93.30          & 94.70           & 49.60 & \uline{96.44}   \scriptsize{$\pm$0.08}  \\
        Fire Hood         & 90.75 & 50.97 & 81.93          & \textbf{97.53} & 95.45          & \uline{96.60}  & 88.32          & 95.60           & 80.68 & 93.80           \scriptsize{$\pm$0.21}  \\
\end{tabular}
    }
    \label{tab:appendix_realiad_s-auroc}
\end{table*}

\begin{table*}
    \centering
    \caption{Comprehensive results for anomaly segmentation, measured as \textbf{AUPRO} ($\uparrow$) on Real-IAD. The best performing method is marked in \textbf{bold} with the runner-up \underline{underlined}. Our method is run $n=5$ times.}
    \resizebox{\linewidth}{!}{
    \begin{tabular}{l||c|c|c|c|c|c|c|c|c|c}
    \multirow{2}{*}{Category}    &  CFA   & DSR   & EfficientAD & FastFlow & Multi-Flow & PaDiM & RD4AD    & SimpleNet & STFPM & \method          \\
    & \cite{cfa} & \cite{zavrtanik2022dsr} & \cite{efficientAD} & \cite{fastflow} & \cite{MultiFlow} & \cite{padim} & \cite{rd} & \cite{simplenet} & \cite{stfpm} & \emph{(ours)} \\\midrule
        Audiojack         & 85.29 & 75.12 & 84.52         & 66.94 & 88.03          & 78.94 & \textbf{96.98} & 70.28         & 80.66         & \uline{90.98}  \scriptsize{$\pm$ 0.05}  \\
        Bottle Cap        & 94.93 & 43.91 & 67.88         & 85.22 & \uline{95.73}  & 87.91 & 91.92          & 89.61         & 86.98         & \textbf{98.49} \scriptsize{$\pm$ 0.01}  \\
        Button Battery    & 58.99 & 37.21 & 44.08         & 30.33 & 90.75          & 51.40 & \uline{93.72}  & 73.28         & 69.84         & \textbf{93.72} \scriptsize{$\pm$ 0.03}  \\
        End Cap           & 82.08 & 62.09 & 85.01         & 63.19 & 82.85          & 84.75 & \textbf{94.71} & 00.03          & 66.47         & \uline{94.22}  \scriptsize{$\pm$ 0.02}  \\
        Eraser            & 84.94 & 50.28 & 76.82         & 78.58 & 92.64          & 83.61 & 89.38          & 86.58         & \uline{95.34} & \textbf{97.24} \scriptsize{$\pm$ 0.04}  \\
        Fire Hood         & 79.66 & 35.96 & 64.72         & 61.79 & \textbf{91.61} & 68.06 & 78.38          & 79.45         & 84.79         & \uline{91.52}  \scriptsize{$\pm$ 0.06}  \\
\end{tabular}
    }
    \label{tab:appendix_realiad_p_aupro}
\end{table*}

\begin{table*}
    \centering
    \caption{Comprehensive results for anomaly segmentation, measured as \textbf{P-AUROC} ($\uparrow$) on Real-IAD. The best performing method is marked in \textbf{bold} with the runner-up \underline{underlined}. Our method is run $n=5$ times.}
    \resizebox{\linewidth}{!}{
    \begin{tabular}{l||c|c|c|c|c|c|c|c|c|c}
    \multirow{2}{*}{Category}    &  CFA   & DSR   & EfficientAD & FastFlow & Multi-Flow & PaDiM & RD4AD    & SimpleNet & STFPM & \method          \\
    & \cite{cfa} & \cite{zavrtanik2022dsr} & \cite{efficientAD} & \cite{fastflow} & \cite{MultiFlow} & \cite{padim} & \cite{rd} & \cite{simplenet} & \cite{stfpm} & \emph{(ours)} \\\midrule
        Audiojack         & 97.62 & 90.31 & 96.36 & 90.56 & 97.68         & 95.17 & \uline{98.62}  & 89.78          & 95.89 & \textbf{99.13} \scriptsize{$\pm$ 0.01}  \\
        Bottle Cap        & 99.28 & 82.34 & 90.94 & 97.04 & 99.08         & 97.97 & \uline{99.65}  & 98.32          & 97.68 & \textbf{99.84} \scriptsize{$\pm$ 0.00}  \\
        Button Battery    & 94.82 & 66.29 & 83.24 & 80.79 & 98.57         & 91.45 & \uline{99.02}  & 97.69          & 91.46 & \textbf{99.20} \scriptsize{$\pm$ 0.01}  \\
        End Cap           & 93.23 & 76.40 & 92.37 & 80.36 & 95.45         & 94.92 & \uline{98.20}  & 28.60          & 88.67 & \textbf{98.46} \scriptsize{$\pm$ 0.01}  \\
        Eraser            & 97.04 & 67.23 & 90.88 & 95.76 & 98.83         & 97.13 & \uline{99.38}  & 97.91          & 99.27 & \textbf{99.67}\scriptsize{$\pm$ 0.00}  \\
        Fire Hood         & 97.25 & 48.70 & 92.79 & 92.43 & 98.35         & 95.19 & \uline{99.00}  & 97.01          & 97.90 & \textbf{99.48}\scriptsize{$\pm$ 0.00}  \\
\end{tabular}
    }
    \label{tab:appendix_realiad_p_auroc}
\end{table*}

\begin{table*}
    \centering
    \caption{Comprehensive results for anomaly segmentation, measured as \textbf{P-AP} ($\uparrow$) on Real-IAD. The best performing method is marked in \textbf{bold} with the runner-up \underline{underlined}. Our method is run $n=5$ times.}
    \resizebox{\linewidth}{!}{
    \begin{tabular}{l||c|c|c|c|c|c|c|c|c|c}
    \multirow{2}{*}{Category}    &  CFA   & DSR   & EfficientAD & FastFlow & Multi-Flow & PaDiM & RD4AD    & SimpleNet & STFPM & \method          \\
    & \cite{cfa} & \cite{zavrtanik2022dsr} & \cite{efficientAD} & \cite{fastflow} & \cite{MultiFlow} & \cite{padim} & \cite{rd} & \cite{simplenet} & \cite{stfpm} & \emph{(ours)} \\\midrule
        Audiojack         & 14.76 & 28.08          & 11.17          & 01.73  & 14.35 & 01.51 & \uline{31.55}  & 11.51          & 30.03         & \textbf{36.02} \scriptsize{$\pm$ 0.22}  \\
        Bottle Cap        & 20.13 & 06.31          & 06.07          & 17.69  & 06.04 & 12.91 & \uline{31.75}  & 14.57          & 07.24         & \textbf{32.50} \scriptsize{$\pm$ 0.07}  \\
        Button Battery    & 10.30 & 12.92          & 25.32          & 02.41  & 23.18 & 05.84 & \textbf{51.10} & 21.94          & 20.37         & \uline{49.14}  \scriptsize{$\pm$ 0.16}  \\
        End Cap           & 07.79 & 07.27          & \textbf{15.61} & 01.83  & 04.17 & 03.52 & 13.12          & 00.04          & 03.66         & \uline{13.60}  \scriptsize{$\pm$ 0.10}   \\
        Eraser            & 18.70 & 20.64          & 10.62          & 06.02  & 08.74 & 05.87 & \textbf{30.40} & 15.88          & 29.32         & \uline{30.25}  \scriptsize{$\pm$ 0.14}  \\
        Fire Hood         & 12.61 & 11.90          & 06.44          & 02.99  & 09.03 & 02.90 & \textbf{31.56} & 09.28          & 20.14         & \uline{27.94}  \scriptsize{$\pm$ 0.20}   \\
\end{tabular}
    }
    \label{tab:appendix_realiad_p_ap}
\end{table*}

\begin{table*}
    \centering
    \caption{Comprehensive results for anomaly detection, measured as \textbf{I-AUROC} ($\uparrow$) on MANTA-Tiny, sorted by category. The best performing method is marked in \textbf{bold} with the runner-up \underline{underlined}. Our method is run $n=5$ times.}
    \resizebox{\linewidth}{!}{
    \begin{tabular}{c|l||c|c|c|c|c|c|c|c|c|c}
    & \multirow{2}{*}{Category}    &  CFA   & DSR   & EfficientAD & FastFlow & Multi-Flow & PaDiM & RD4AD    & SimpleNet & STFPM & \method          \\
    & & \cite{cfa} & \cite{zavrtanik2022dsr} & \cite{efficientAD} & \cite{fastflow} & \cite{MultiFlow} & \cite{padim} & \cite{rd} & \cite{simplenet} & \cite{stfpm} & \emph{(ours)} \\\toprule
        \multirow{4}{*}{\rotatebox[origin=c]{90}{Agriculture}} &Maize                 & 59.98         & 43.06 & 62.92          & 66.19 & 66.97         & 50.90 & \uline{72.65}  & \textbf{88.31} & 64.19 & 64.62           \scriptsize{$\pm$0.18}  \\
        &Paddy                 & 51.00         & 75.15 & \textbf{91.91} & 72.56 & \uline{80.98} & 52.09 & 78.50          & 74.02          & 70.34 & 80.48           \scriptsize{$\pm$0.27}  \\
        &Soybean               & 74.96         & 68.49 & \uline{91.77}  & 85.04 & 83.97         & 59.01 & 88.72          & \textbf{94.44} & 82.69 & 85.06           \scriptsize{$\pm$0.16}  \\
        &Wheat                 & 50.11         & 76.16 & 83.94          & 66.68 & 81.32         & 53.24 & 84.56          & \textbf{93.49} & 66.77 & \uline{86.92}   \scriptsize{$\pm$0.16}  \\\midrule
        \multirow{11}{*}{\rotatebox[origin=c]{90}{Electronics}}&Block Inductor       & 76.29         & 64.42 & \textbf{96.86} & 84.04 & 87.16         & 62.48 & 90.89          & 91.45          & 84.76 & \uline{92.18}   \scriptsize{$\pm$0.16}  \\
        &Copper Standoff      & 68.23         & 90.38 & 96.72          & 85.61 & 92.20         & 58.45 & \uline{97.08}  & 95.39          & 68.73 & \textbf{97.27}  \scriptsize{$\pm$0.13}  \\
\end{tabular}
    }
    \label{tab:appendix_manta_i-auroc}
\end{table*}

\begin{table*}
    \centering
    \caption{Comprehensive results for anomaly detection, measured as \textbf{S-AUROC} ($\uparrow$) on MANTA-Tiny, sorted by category. The best performing method is marked in \textbf{bold} with the runner-up \underline{underlined}. Our method is run $n=5$ times.}
    \resizebox{\linewidth}{!}{
    \begin{tabular}{c|l||c|c|c|c|c|c|c|c|c|c}
    & \multirow{2}{*}{Category}    &  CFA   & DSR   & EfficientAD & FastFlow & Multi-Flow & PaDiM & RD4AD    & SimpleNet & STFPM & \method          \\
    & & \cite{cfa} & \cite{zavrtanik2022dsr} & \cite{efficientAD} & \cite{fastflow} & \cite{MultiFlow} & \cite{padim} & \cite{rd} & \cite{simplenet} & \cite{stfpm} & \emph{(ours)} \\\toprule
        \multirow{4}{*}{\rotatebox[origin=c]{90}{Agriculture}} & Maize               & 52.96 & 33.91 & 67.81          & 62.82         & 70.53         & 63.37 & \uline{75.53}  & \textbf{90.38} & 62.22 & 60.64           \scriptsize{$\pm$0.38}  \\
         & Paddy               & 57.73 & 65.50 & 88.75          & 68.78         & \uline{88.86} & 80.89 & 79.02          & \textbf{95.55} & 77.39 & 83.12           \scriptsize{$\pm$0.53}  \\
         & Soybean             & 75.52 & 65.80 & 90.23          & 85.54         & \uline{87.19} & 85.74 & 85.46          & \textbf{93.71} & 83.71 & 80.56           \scriptsize{$\pm$0.35}  \\
         & Wheat               & 52.33 & 81.52 & 84.91          & 68.52         & 85.18         & 80.63 & \uline{86.93}  & \textbf{94.75} & 62.28 & 83.80           \scriptsize{$\pm$0.58}  \\\midrule
        \multirow{11}{*}{\rotatebox[origin=c]{90}{Electronics}} & Block Inductor      & 71.45 & 54.44 & \uline{91.12}  & 78.10         & 86.67         & 76.78 & 90.50          & \textbf{91.26} & 74.55 & 88.33           \scriptsize{$\pm$0.25}  \\
         & Copper Standoff     & 74.35 & 86.92 & 93.63          & 72.37         & 95.35         & 90.90 & \uline{97.84}  & 95.8           & 92.14 & \textbf{98.25}  \scriptsize{$\pm$0.13}  \\
\end{tabular}
    }
    \label{tab:appendix_manta_s-auroc}
\end{table*}

\begin{table*}
    \centering
    \caption{Comprehensive results for anomaly segmentation, measured as \textbf{AUPRO} ($\uparrow$) on MANTA-Tiny, sorted by category. The best performing method is marked in \textbf{bold} with the runner-up \underline{underlined}. Our method is run $n=5$ times.}
   \resizebox{\linewidth}{!}{
    \begin{tabular}{c|l||c|c|c|c|c|c|c|c|c|c}
    & \multirow{2}{*}{Category}    &  CFA   & DSR   & EfficientAD & FastFlow & Multi-Flow & PaDiM & RD4AD    & SimpleNet & STFPM & \method          \\
    & & \cite{cfa} & \cite{zavrtanik2022dsr} & \cite{efficientAD} & \cite{fastflow} & \cite{MultiFlow} & \cite{padim} & \cite{rd} & \cite{simplenet} & \cite{stfpm} & \emph{(ours)} \\\toprule
        \multirow{4}{*}{\rotatebox[origin=c]{90}{Agriculture}} & Maize                 & 38.99 & 48.53 & 44.13          & 40.49 & 62.33         & 55.29 & \uline{65.69}  & 61.67         & 40.15 &\textbf{68.32} \scriptsize{$\pm$  0.12}  \\
        &Paddy                 & 41.22 & 48.42 & 63.83          & 52.66 & 64.57         & 59.19 & \uline{65.27}  & 39.12         & 55.55 & \textbf{71.41} \scriptsize{$\pm$  0.09}  \\
        &Soybean               & 50.61 & 37.76 & 53.31          & 52.07 & 73.44         & 66.83 & \uline{74.25}  & 67.62         & 51.43 & \textbf{78.04} \scriptsize{$\pm$  0.04}  \\
        &Wheat                 & 41.47 & 65.56 & 58.40          & 48.11 & \uline{77.31} & 72.95 & 73.76          & 71.28         & 45.74 & \textbf{84.33} \scriptsize{$\pm$  0.05}  \\\midrule
        \multirow{11}{*}{\rotatebox[origin=c]{90}{Electronics}} &Block Inductor       & 60.58 & 60.30 & 86.94          & 68.38 & 81.40          & 82.78 & \uline{89.86}  & 80.35         & 68.37 & \textbf{92.60} \scriptsize{$\pm$  0.07}  \\
        &Copper Standoff      & 37.17 & 43.98 & \uline{92.26}  & 71.47 & 85.01          & 83.77 & 92.09          & 80.74         & 31.15 & \textbf{93.72} \scriptsize{$\pm$  0.04}  \\
\end{tabular}
   }
    \label{tab:appendix_manta_p-aupro}
\end{table*}

\begin{table*}
    \centering
    \caption{Comprehensive results for anomaly segmentation, measured as \textbf{P-AUROC} ($\uparrow$) on MANTA-Tiny, sorted by category. The best performing method is marked in \textbf{bold} with the runner-up \underline{underlined}. Our method is run $n=5$ times.}
   \resizebox{\linewidth}{!}{
    \begin{tabular}{c|l||c|c|c|c|c|c|c|c|c|c}
    & \multirow{2}{*}{Category}    &  CFA   & DSR   & EfficientAD & FastFlow & Multi-Flow & PaDiM & RD4AD    & SimpleNet & STFPM & \method          \\
    & & \cite{cfa} & \cite{zavrtanik2022dsr} & \cite{efficientAD} & \cite{fastflow} & \cite{MultiFlow} & \cite{padim} & \cite{rd} & \cite{simplenet} & \cite{stfpm} & \emph{(ours)} \\\toprule
        \multirow{4}{*}{\rotatebox[origin=c]{90}{Agriculture}}&Maize               & 70.14         & 63.32 & 78.86          & 70.81 & 83.04         & 81.03 & \textbf{89.17} & 87.18          & 72.81         & \uline{87.64}   \scriptsize{$\pm$0.07}  \\
        &Paddy               & 66.22         & 50.62 & \textbf{89.68} & 63.29 & \uline{77.00} & 71.85 & 75.57          & 80.13          & 73.24         & 76.71           \scriptsize{$\pm$0.04}  \\
        &Soybean             & 69.06         & 56.09 & 85.05          & 72.60 & 88.25         & 87.53 & \textbf{90.99} & 88.93          & 80.46         & \uline{90.20}   \scriptsize{$\pm$0.03}  \\
        &Wheat               & 77.97         & 84.90 & 84.42          & 79.96 & 91.39         & 93.01 & \uline{93.14}  & 92.53          & 79.88         & \textbf{95.44}  \scriptsize{$\pm$0.01}  \\\midrule
        \multirow{11}{*}{\rotatebox[origin=c]{90}{Electronics}} &Block Inductor      & 91.54         & 78.54 & 96.16          & 90.79 & 95.17         & 94.83 & \uline{97.81}  & 96.90          & 90.71         & \textbf{98.06}  \scriptsize{$\pm$0.01}  \\
        &Copper Standoff     & 78.05         & 74.82 & \textbf{98.82} & 92.54 & 95.04         & 95.98 & 98.29          & 95.68          & 71.77         & \uline{98.74}   \scriptsize{$\pm$0.01}  \\
\end{tabular}
   }
    \label{tab:appendix_manta_p-auroc}
\end{table*}

\begin{table*}
    \centering
    \caption{Comprehensive results for anomaly segmentation, measured as \textbf{P-AP} ($\uparrow$) on MANTA-Tiny, sorted by category. The best performing method is marked in \textbf{bold} with the runner-up \underline{underlined}. Our method is run $n=5$ times.}
   \resizebox{\linewidth}{!}{
    \begin{tabular}{c|l||c|c|c|c|c|c|c|c|c|c}
    & \multirow{2}{*}{Category}    &  CFA   & DSR   & EfficientAD & FastFlow & Multi-Flow & PaDiM & RD4AD    & SimpleNet & STFPM & \method          \\
    & & \cite{cfa} & \cite{zavrtanik2022dsr} & \cite{efficientAD} & \cite{fastflow} & \cite{MultiFlow} & \cite{padim} & \cite{rd} & \cite{simplenet} & \cite{stfpm} & \emph{(ours)} \\\toprule
        \multirow{4}{*}{\rotatebox[origin=c]{90}{Agriculture}}&Maize               & 08.88 & 09.44 & 28.05          & 11.84         & 16.20          & 21.04          & \uline{29.67}  & \textbf{33.72} & 12.10         & 19.68           \scriptsize{$\pm$0.12}  \\
        &Paddy               & 06.26 & 08.22 & \textbf{32.96} & 08.62         & 12.46          & \uline{16.79}  & 14.89          & 20.11          & 10.22         & 12.94           \scriptsize{$\pm$0.02}  \\
        &Soybean             & 18.36 & 13.83 & 35.45          & 21.16         & 36.80          & \uline{45.92}  & \textbf{47.16} & 45.16          & 28.46         & 38.86           \scriptsize{$\pm$0.06}  \\
        &Wheat               & 04.83 & 21.96 & 15.63          & 08.80         & 27.65          & \uline{36.09}  & 30.82          & 43.89          & 07.91         & \textbf{36.63}  \scriptsize{$\pm$0.13}  \\\midrule
        \multirow{11}{*}{\rotatebox[origin=c]{90}{Electronics}}&Block Inductor      & 11.69 & 16.11 & 40.24          & 11.51         & 43.03          & 41.81          & \textbf{56.03} & 44.51          & 33.36         & \uline{53.44}   \scriptsize{$\pm$0.06}  \\
        &Copper Standoff     & 03.47 & 11.69 & 56.63          & 17.58         & 38.95          & 44.98          & \textbf{62.85} & 41.28          & 03.11         & \uline{60.47}   \scriptsize{$\pm$0.18}  \\
\end{tabular}
   }
    \label{tab:appendix_manta_p-ap}
\end{table*}